\documentclass[10pt,twocolumn,letterpaper]{article}

\usepackage[pagenumbers]{cvpr} 

\usepackage{multirow}
\usepackage{makecell}
\usepackage{adjustbox}
\usepackage{amsmath,amssymb}
\usepackage{algorithm}
\usepackage{algpseudocode}
\usepackage{xr-hyper}
\DeclareRobustCommand{\claude}[1]{{#1}}

\newcommand{\bench}{CogCanvas\xspace}
\newcommand{\method}{Context2Canvas\xspace}

\definecolor{cvprblue}{rgb}{0.21,0.49,0.74}
\usepackage[pagebackref,breaklinks,colorlinks,allcolors=cvprblue]{hyperref}

\DeclareRobustCommand{\revise}[1]{#1}

\usepackage{colortbl}
\definecolor{oursbg}{HTML}{D6E6F7}
\newcommand{\best}[1]{\cellcolor{red!25}\textbf{#1}}
\newcommand{\second}[1]{\cellcolor{orange!25}#1}
\newcommand{\third}[1]{\cellcolor{yellow!25}#1}
\newcommand{\oursc}[1]{\cellcolor{oursbg}#1}
\DeclareRobustCommand{\legbox}[2]{{\setlength{\fboxsep}{1pt}\colorbox{#1}{\strut #2}}}
\newcommand{\tablelegend}{Within each group, the best result is shaded \legbox{red!25}{\textbf{red}}, the second best \legbox{orange!25}{orange}, and the third best \legbox{yellow!25}{yellow}; rows shaded \legbox{oursbg}{blue} are our method.}

\usepackage{microtype} 

\usepackage{cuted}        
\usepackage{multirow}
\usepackage{makecell}
\usepackage{adjustbox}
\usepackage{amsmath,amssymb}
\usepackage{pifont}       
\usepackage{orcidlink}
\makeatletter\let\@author\@empty\makeatother 
\usepackage{authblk}

\def\confName{CVPR}
\def\confYear{2027}

\title{\claude{From Context to Canvas: Benchmark Dataset and Verify-then-Route Framework\\for Multi-Subject Reference-Based Image Generation}}

\let\ABorigrlap\rlap
\DeclareRobustCommand{\projectpage}{{\let\rlap\ABorigrlap\href{https://longbaocoder2.github.io/cogcanvas}{{\setlength{\fboxsep}{1.5pt}\colorbox[HTML]{FDF0F4}{\strut\textcolor[HTML]{D6336C}{\texttt{https://longbaocoder2.github.io/cogcanvas}}}}}}}

\author[1,2$\dagger$\orcidlink{0009-0003-6311-745X}]{Long-Bao Nguyen}
\author[1,2$\dagger$\orcidlink{0009-0004-3733-6496}]{Quang-Khai Le}
\author[3\orcidlink{0000-0003-0236-7992}]{Tam V. Nguyen}
\author[1,2\orcidlink{0000-0003-3046-3041}]{\protect\\Minh-Triet Tran}
\author[1,2$\ddagger$\orcidlink{0000-0002-7363-2610}]{Trung-Nghia Le}
\affil[1]{University of Science, Ho Chi Minh City, Vietnam}
\affil[2]{Vietnam National University, Ho Chi Minh City, Vietnam}
\affil[3]{University of Dayton, Ohio, United States}

\affil[ ]{\textbf{Project page}:~\projectpage}

\begin{document}

\twocolumn[{%
\renewcommand\twocolumn[1][]{#1}%
\maketitle
\begin{center}
  \centering
  \includegraphics[clip,width=\linewidth]{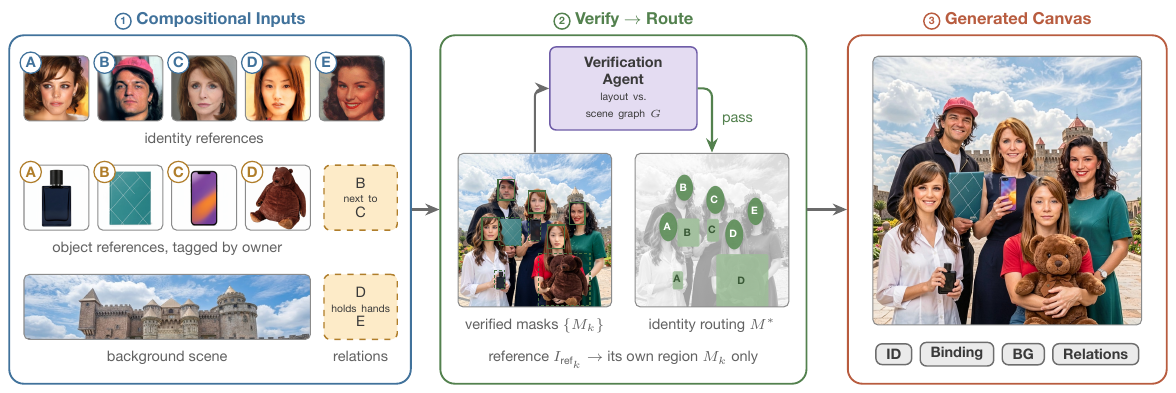}
  \vspace{-5mm}
  \captionof{figure}{\claude{\textbf{Multi-subject reference-based generation as studied in this paper.} Given $N\!\in\!\{2,\dots,5\}$ identity references, per-person object and fashion references, interactions, and a background reference, a generator must return one image that preserves every identity, binds each item to the right person, follows the layout, and keeps the scene. \textbf{\bench} turns each requirement into a checkable assertion scored on six axes; \textbf{\method} verifies a layout against a scene graph before injecting any identity, then routes each reference only to its own verified region.}}
  \label{fig:teaser}
\end{center}
}]

{\renewcommand{\thefootnote}{}\footnotetext{$^{\dagger}$Equal contribution.\quad $^{\ddagger}$Corresponding author.}}

\begin{abstract}
\revise{Personalized image generation should compose several people, each with the right objects and outfits, in a specified scene, yet current generators collapse as references are added and existing benchmarks do not capture this degradation. We introduce CogCanvas, a benchmark of 1,361 compositional prompts with 2--5 subjects, 1,952 curated references (100 people, 115 objects and fashion items, 29 scenes), and graphs specifying interactions and spatial relations. It adds two metrics: BG-Sim for background fidelity outside foreground regions, and Attr-VQA for person--item binding, with 76.5\% agreement with human majority judgments. We also propose Context2Canvas, a training-free multi-agent framework built on a verify-then-route principle: agents verify a layout against a hierarchical scene graph, and its masks then route each reference to its own region in a two-phase renderer that grounds references independently before refining them jointly. Context2Canvas achieves 2.5$\times$ the attribute-binding score of the strongest published baseline, with larger gains as the subject count increases.}

\end{abstract}

\section{Introduction}
\label{sec:intro}

\claude{Diffusion and flow-based generators have made text-to-image synthesis remarkably faithful, progressing from U-Net models~\cite{Nichol2021GLIDETP,rombach2022high} to multimodal diffusion transformers~\cite{3692070.3692573,flux2024}.} \emph{Personalized} generation extends this capability to specific subjects: given reference images, the model must reproduce each subject's identity and appearance while following the prompt. \claude{Reference-conditioned systems~\cite{labs2025flux1kontextflowmatching,xiao2024omnigen,omnigen2,wu2025uno} now perform well when one subject is given.}


Group travel photos and visual storytelling call for composing several people and their objects in a specified scene. For example, “At ⟨landmark⟩, A holds a perfume bottle on the left while B and C hold hands on the right.” The image must preserve identities, assign objects correctly, and follow the specified positions, interactions, and background. \revise{Yet current methods struggle as reference inputs grow: the strongest published method's face-identity score drops 81-fold from two to five subjects (\cref{sec:exp:main}). Two gaps hinder progress, and we address both.}

\claude{
Existing benchmarks test one requirement at a time (\cref{tab:benchmark_comparison}): compositional benchmarks~\cite{ghosh2023geneval,huang2023t2icompbench} have no identity references, personalization benchmarks~\cite{ruiz2023dreambooth,peng2025dreambenchpp} are single-subject, and multi-subject benchmarks~\cite{multihuman2025,xverse2025} have no background reference and do not test who holds what. Their global scores also hide the failure that matters: an image that has lost every identity still earns 83\% of the best CLIP-T score. 
When several references share one denoising pass, nothing tells attention which region belongs to which subject, so identities blend and attributes migrate; layout-conditioned methods supply regions but apply them open-loop, so a missing subject propagates to the output.}

\claude{We introduce \textbf{\bench}, which places 2--5 of 100 identities, each bound to one of 115 distinctive objects or fashion items, in one of 29 recognizable scenes, and reduces each of its 1,361 prompts to interaction and position graphs so that every relation becomes a separately checkable question. We then propose \textbf{\method}, a training-free multi-agent framework built on a \emph{verify-then-route} principle (Fig.~\ref{fig:teaser}): agents compile the prompt into a hierarchical scene graph and verify a generated layout against it in a closed loop, and the verified masks define a deterministic attention routing mask for a two-phase renderer that grounds each reference in isolation before harmonizing all subjects jointly.}

\claude{Our contributions are as follows:
\begin{itemize}[nosep,leftmargin=*]
  \item We introduce \textbf{\bench}, a benchmark that jointly covers human identity, object/fashion, and background references with graph-structured ground truth, plus two metrics, BG-Sim and Attr-VQA.
  \item We propose \textbf{\method}, a training-free framework whose scene-graph-verified layout directly supplies the spatial condition for identity injection, rendered by a \textit{two-phase renderer with a deterministic routing mask}.
  \item We conduct experiments demonstrating that state-of-the-art generators collapse as subject count increases, while global metrics fail to capture this degradation. \method improves attribute binding by $2.5\times$ over the strongest published baseline.
\end{itemize}}

\begin{figure*}[t]
  \centering
  \includegraphics[clip,width=\linewidth]{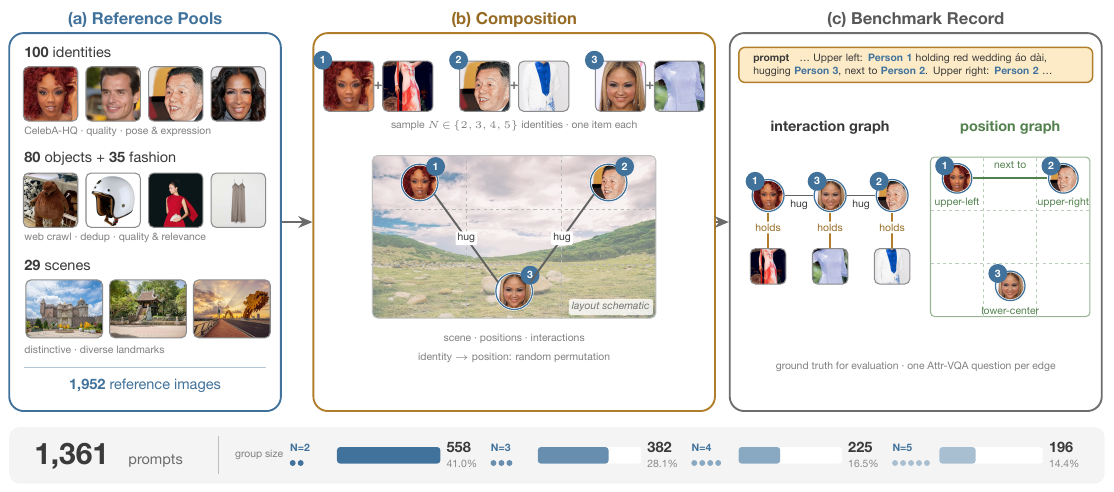}
  \vspace{-6mm}
  \caption{\textbf{\bench data curation pipeline.}
  \emph{Left:} Three disjoint reference pools are curated independently: person identities from CelebA-HQ, object/fashion items via web crawling with DINOv2 deduplication and two-stage filtering, and background scenes selected for global diversity.
  \emph{Center:} Compositional prompts are constructed by sampling $N\in\{2,\dots,5\}$ identities, assigning per-person object/fashion references, selecting a background scene, defining pairwise interactions, and \claude{assigning explicit positions sampled over the $N!$ orderings of the group}.
  \emph{Right:} The resulting 1,361 prompts follow a long-tail group-size distribution and are annotated with structured interaction and position graphs that serve as ground-truth supervision for attribute-binding and spatial-plausibility evaluation.}
  \label{fig:curation}
  \vspace{-3mm}
\end{figure*}

\section{Related Work}
\label{sec:related}

\begin{table}[t]
  \centering
  \caption{\claude{\textbf{Benchmark comparison.} \#Prompts: evaluation prompts. Human/Object: number of identity and object references. HOI: explicit human--object interaction evaluation. BG: background scene reference provided. Spatial: spatial relations between subjects evaluated. Subject counts are detailed in Supp.~A.3. \revise{Our benchmark is shaded \legbox{oursbg}{blue}.}}}
  \label{tab:benchmark_comparison}
  \footnotesize
  \setlength{\tabcolsep}{3pt}
  \begin{adjustbox}{max width=\linewidth}
  \begin{tabular}{@{}l l r r r c c c@{}}
    \toprule
    \textbf{Benchmark} & \textbf{Venue} & \textbf{\#Prompts}
      & \textbf{Human} & \textbf{Object} & \textbf{HOI} & \textbf{BG}$\uparrow$ & \textbf{Spatial} \\
    \midrule
    T2I-CompBench~\cite{huang2023t2icompbench} & NeurIPS'23 & 6{,}000 & 0 & 0 & \ding{55} & \ding{55} & $\circ$ \\
    DreamBench~\cite{ruiz2023dreambooth} & CVPR'23 & 750 & 0 & 30 & \ding{55} & \ding{55} & \ding{55} \\
    DreamBench++~\cite{peng2025dreambenchpp} & ICLR'25 & 1{,}350 & 20 & 110 & \ding{55} & \ding{55} & \ding{55} \\
    MultiHuman-TB~\cite{multihuman2025} & NeurIPS'25 & 1{,}800 & 5{,}550 & 0 & $\circ$ & \ding{55} & \ding{55} \\
    XVerseBench~\cite{xverse2025} & NeurIPS'25 & 300 & 20 & 119 & $\circ$ & \ding{55} & \ding{55} \\
    \oursc{\textbf{\bench (ours)}} & \oursc{--} & \oursc{1{,}361} & \oursc{100} & \oursc{115} & \oursc{\ding{51}} & \oursc{\ding{51}} & \oursc{\ding{51}} \\
    \bottomrule
  \end{tabular}
  \end{adjustbox}
  \vspace{1pt}

  {\scriptsize \claude{\ding{51}: full; $\circ$: partial; \ding{55}: none.}}
  \vspace{-3mm}
\end{table}

\claude{\textbf{Multi-subject personalization.}} Personalized generation reproduces user-specified subjects while following the prompt. Methods fine-tune per subject~\cite{gal2022textualinversion,ruiz2023dreambooth}, train image-conditioned adapters~\cite{ye2023ipadapter,tan2024ominicontrol}, or compose several references in one model~\cite{xiao2024omnigen,omnigen2,xverse2025,wu2025uno,dreamo2025,mosaic2025}. \claude{In all of them identity fidelity degrades as subjects are added, because references compete within shared attention and leak features into one another. TF-TI2I~\cite{Hsiao2025TFTI2ITT} routes references inside joint attention, but assigns each output token online from attention scores, which can oscillate between visually similar subjects; MS-Diffusion~\cite{wang2024msdiffusion} restricts reference attention to regions but takes the layout as a user-supplied, unchecked input. Our routing assignment is fixed by a verified layout before sampling begins.}

\claude{\textbf{Compositional and agentic generation.}} Layout-conditioned methods impose structure through boxes, layouts, or masked attention~\cite{wang2024msdiffusion,lmd2023,rpg2024}, but they are open-loop: the layout is applied once and never checked, so a missing object or a wrong containment relation propagates to the final image. Agentic methods close this loop with LLM or VLM feedback, improving compositional binding~\cite{mccd2025}, instance counts~\cite{countloop2025}, and foreground consistency~\cite{anywhere2024}. \claude{\revise{Most of them} check the image against a flat description of the prompt and do not exploit part--whole structure, which is what makes small, held, or occluded objects findable. \revise{LayoutAgent~\cite{fan2025layoutagent} does build a scene graph with a VLM agent, but uses it to plan object boxes before generation and conditions on object images, not identity references.} We verify against a hierarchical scene graph and search for each child inside its parent's region first\revise{, and the verified masks then fix how identity references are routed during rendering, which no agentic pipeline above controls}.}

\revise{\textbf{Learned routing and unified canvases.} Recent methods also decouple \emph{where} a reference acts from \emph{what} it contributes, but learn this decoupling: MOSAIC~\cite{mosaic2025} supervises correspondence with disentanglement losses, RefRoute~\cite{he2026refroute} routes compact reference tokens by an attention mask after many-reference fine-tuning, and RefCompose~\cite{kuppa2026refcompose} and Canvas-to-Image~\cite{dalva2025canvas} pack references into one fixed-size canvas at constant cost. \method trains nothing and fixes each token's routing from a layout verified before sampling, trading latency for a hard, inspectable assignment; we see it as complementary to these methods (Supp.~D.3).}


\claude{\textbf{Benchmarks.} Preference and compositional benchmarks~\cite{wu2023hpd,ghosh2023geneval,huang2023t2icompbench,evalmuse40k} score attributes and pairwise relations between generic objects without identity references; DreamBench(++)~\cite{ruiz2023dreambooth,peng2025dreambenchpp} target single subjects, and XVerseBench~\cite{xverse2025} reaches three subjects without testing interactions or layouts. MultiHuman-Testbench~\cite{multihuman2025}, the closest to our setting, scores face count, identity, and simple group actions, but has no object or background references and no per-person binding (\cref{tab:benchmark_comparison}).}

\section{\bench Benchmark}
\label{sec:bench}

\subsection{Task Definition}
\label{sec:bench:task}

\claude{The primary task is \textbf{multi-subject reference-based generation}. Each prompt is the tuple $\bigl(B,\{H_i\}_{i=1}^{N},\{O_j\}_{j=1}^{M},\{R_j\}_{j=1}^{M}\bigr)$ with $M\le N$, where $B$ is a background reference, $\{H_i\}$ are $N$ identity references, $\{O_j\}$ are $M$ object or fashion references each bound to a single person, and $\{R_j\}$ are the corresponding human--object interaction specifications, rendered as a natural-language prompt $P$. The schema reduces every prompt to a verifiable set of bindings rather than free-form text. \revise{The data also support text-only compositional generation and other auxiliary tasks (Supp.~A.4).}}

\subsection{Dataset Construction}
\label{sec:bench:construction}

Figure~\ref{fig:curation} shows our data curation pipeline. We curate 1,952 reference images covering 100 identities from CelebA-HQ, 115 objects and fashion items, and 29 recognizable real-world scenes. Portraits vary in pose, expression, lighting, and apparel. Items and scenes are visually distinctive, so matching requires preserving the supplied reference’s appearance. Automated filtering and manual review ensure image quality, while a demographic audit improves balance within the identity pool. Curation criteria and demographic limitations are detailed in the supplementary material.

An MLLM extracts reference attributes and generates prompts that place people in a scene with their assigned items. We sample uniformly across object size and interaction strength to vary compositional difficulty. Each prompt is checked for plausible interactions, feasible layouts, and scene compatibility, then regenerated if it fails. Human overrides are required for 1.2\% of candidates.

Each prompt includes an interaction graph describing person–person and person–object relations and a position graph describing relative subject locations. These graphs are generated and verified by MLLMs. Each edge yields a subject-linked Attr-VQA question, enabling evaluation of specific people and relations. The final dataset contains 1,361 prompts with 2–5 people. Reported averages weight group sizes by their dataset frequencies; detailed statistics appear in the supplementary material.

\begin{figure*}[t]
  \centering
  \includegraphics[width=\textwidth]{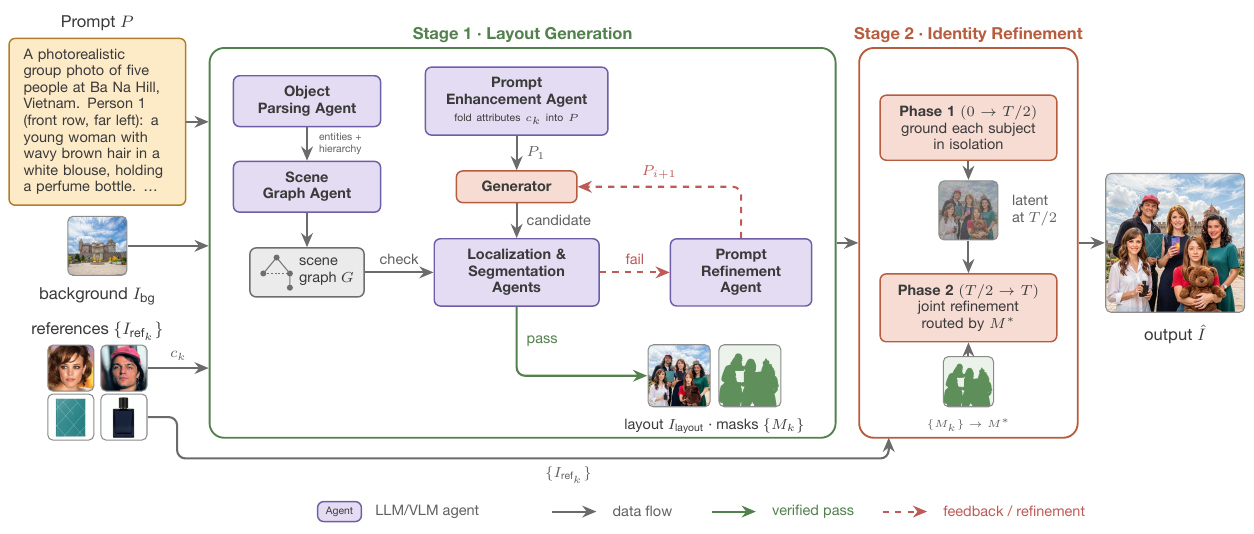}
  \vspace{-5mm}
  \caption{\textbf{Overview of \method.} \revise{Two stages orchestrated by LLM/VLM agents (purple), preceded by an optional retrieval module;} dashed red arrows mark agentic feedback loops. \emph{Layout Generation} (\cref{sec:method:layout}) renders a layout and verifies it against the scene graph, refining the prompt on failure; \emph{Identity Refinement} (\cref{sec:method:refine}) composites all subjects via two-phase denoising with a spatial attention routing mask. No component is fine-tuned at any stage. \emph{Reference Retrieval} (Supp.~B.1) resolves each prompt entity to one reference through a coarse-to-fine funnel\revise{; it is not used in the main experiments, which supply the ground-truth references of \bench to every method}.}
  \label{fig:overview}
  \vspace{-3mm}
\end{figure*}

\subsection{Evaluation Protocol}
\label{sec:bench:metrics}

\claude{Each image is scored on six axes, four standard and two new (details in Supp.~A.4).}



\noindent\textbf{Standard metrics.} ID-Sim measures ArcFace cosine similarity between aligned faces, scaled to [0, 100]. Each reference is matched to its most similar generated face and receives zero if no face is detected\revise{; this best-match rule is not one-to-one, so two references may share a face (Supp.~A.4)}. DINO-Sim similarly matches grounded object and fashion crops to their references using DINOv2\revise{, with the same rule}. CLIP-T and Aesth. measure global text–image alignment and aesthetic quality, respectively, but do not assess identity assignment.

\noindent\textbf{BG-Sim (new).} \claude{BG-Sim measures background fidelity. SAM~3~\cite{carion2025sam3} masks all foreground subjects in both the generated image and the reference background, and we report the cosine similarity of DINOv3~\cite{simeoni2025dinov3} features over the remaining regions. Unlike CLIP-T, it is not dominated by salient subjects, so it directly penalizes a studio backdrop that replaces the requested landmark.}

\noindent\textbf{Attr-VQA (new).} \claude{Attr-VQA asks an MLLM judge (Gemma~4~\cite{gemma2026gemma4}) one binary question per subject--attribute entry and one per subject-pair relation in the interaction and position graphs, \eg, ``Does the person at the lower left hold a perfume bottle?'' or ``Is person $A$ holding hands with person $B$?''. Because every question asserts a ground-truth relation, the score is the percentage of questions answered \emph{yes}, averaged over questions, entities, and prompts. It covers attribute binding, positions, and interactions; it does not test lighting or physical coherence, for which the graphs provide no ground truth.} \claude{We validate Attr-VQA against human judgments in \cref{sec:exp:vqa}.}

\noindent\textbf{Auxiliary tasks.} \claude{For text-only generation we score person-count accuracy and Attr-VQA restricted to binding questions.}

\section{\method Framework}
\label{sec:method}


\method maps a prompt $P$, a background image $I_{bg}$, and a set of reference images $\{(r_k, I_{ref_k})\}_{k=1}^{K}$, each assigned to a prompt entity $r_k$, to an image $\hat{I}$ in which every prompted subject keeps its reference identity and occupies its intended place (\cref{fig:overview}). Single-pass generators misplace subjects and let their references interfere; we address both by verifying the layout first and then using its per-subject masks to constrain identity injection. The framework comprises two stages: Stage~1 (\emph{Layout Generation}) produces a verified layout $I_{layout}$ and masks $\{M_k\}_{k=1}^{K}$ aligned with the supplied references; Stage~2 (\emph{Identity Refinement}) uses these masks to route reference features to their assigned regions and renders $\hat{I}$ through two-phase denoising. Every agent and its backing model are listed in Supp.~B.7.



\subsection{Stage 1: Verified Compositional Layout}
\label{sec:method:layout}

\claude{Single-shot generation often misses entities in crowded prompts because the generator has no explicit representation of scene structure. Stage~1 therefore wraps it in a self-correcting agent loop (Supp.~B.4) that detects missing entities and repairs the prompt.}

\noindent\textbf{Hierarchical scene graph.} Given $P$ and $I_{bg}$, the \textbf{Object Parsing Agent} lists all entities and their containment hierarchy. The hierarchy matters for localization: many objects are defined relative to a parent (a bag is held in a hand, which belongs to a person), so searching for a child inside its parent's region first reduces false negatives without a full-image search. The \textbf{Scene Graph Agent} turns this list into a directed graph $G=(\mathcal{V},\mathcal{E})$ whose nodes carry an entity type, name, and spatial label, and whose edges encode \emph{containment}, \emph{adjacency}, or \emph{interaction}.

\noindent\textbf{Layout generation.} A \textbf{Prompt Enhancement Agent} first folds each entity's attribute description $c_k$ into $P$, yielding $P_1$. Given $I_{bg}$ and the current prompt $P_i$, the generator produces a candidate layout $I_{layout}$ in which the $K$ subjects appear as \emph{anonymous} placeholders arranged according to the prompt's spatial instructions.

\noindent\textbf{Closed-loop verification.} The \textbf{Localization Agent} searches for each node $v$ along the parent--child structure: it first looks inside the region of interest of $v$'s parent and, if the detection confidence stays below a threshold $\theta$, widens the region to the next ancestor (hand~$\to$~person~$\to$~full image), which recovers small and occluded objects. \claude{Let $\mathrm{Anc}(v)=(\pi(v),\pi^{2}(v),\ldots,I)$ denote this chain of regions. Detections are assigned to graph nodes by category-gated Hungarian matching~\cite{kuhn1955hungarian} with an IoU cost\revise{ (this one-to-one matching serves layout verification only and is distinct from the best-match rule used for evaluation in \cref{sec:bench:metrics})}, and the \textbf{Segmentation Agent} produces a binary mask $M_v$ for each localized node.} A node passes when
\begin{equation}
  \mathrm{pass}(v) \;=\; \mathbf{1}\Bigl[\,\max_{R\in\mathrm{Anc}(v)} s_R(v) \,\ge\, \theta \;\wedge\; |M_v|>0\,\Bigr],
  \label{eq:verify}
\end{equation}
\claude{where $s_R(v)$ is the detector confidence for $v$ inside region $R$, and the layout passes when every $v\in\mathcal{V}$ passes. Failing nodes form a structured report (name, parent, and whether detection or segmentation failed) that the \textbf{Prompt Refinement Agent} uses to rewrite $P_i$ into $P_{i+1}$, making those entities' position and appearance explicit. The loop ends when all nodes pass or after $T_{\max}$ iterations, in which case the iteration with the most passing nodes is used.}


\noindent\textbf{What is and is not verified.}
At this stage, reference images have not yet been introduced, so verification checks whether the base image contains the entity categories required for subsequent generation (e.g., a book or a perfume bottle), rather than whether their appearance matches the references. Spatial and interaction relations are specified in the prompt but are not explicitly verified in the generation loop; their fidelity is evaluated afterward using Attr-VQA. A layout containing all required categories can therefore pass despite incorrect or physically implausible relations (\cref{sec:exp:qual}).

\subsection{Stage 2: Two-Phase Identity Refinement}
\label{sec:method:refine}

Stage~2 renders the verified layout into the final image $\hat{I}$ without any training. \claude{Placing $K$ identities on one canvas poses two competing demands: each subject must land in its own region, yet injecting reference features region by region shifts the latent distribution and leaves seams and truncated identity detail. We therefore split the $T$-step denoising trajectory, indexed by $t$ from pure noise ($t{=}0$) to the final latent ($t{=}T$), into two phases (\cref{fig:overview}, right; detailed in Supp.~B.2).}

\noindent\textbf{Encoding.} A shared VAE encoder $\mathcal{E}_i$ maps the layout and each reference to latent tokens, $\bar{\tau}_{I_0}=\mathcal{E}_i(I_0)$ with $I_0:=I_{layout}$ and $\tau_{R_k}=\mathcal{E}_i(I_{ref_k})$; a text encoder $\mathcal{E}_p$ maps the per-subject prompts and the global scene prompt to $\tau_{P_k}$ and $\tau_{P_0}$. 
Masks are not encoded: each $M_k$ is smoothed with a Gaussian blur, binarized at output resolution, and resampled onto the token grid by nearest-neighbour interpolation (Supp.~B.3), so mask boundaries coincide with the tokens that attention sees.
Only the canvas latent $\tau^{t}_{I_0}$ carries the step index. Following the RoPE~\cite{su2024roformer} offset design of FLUX.2~\cite{bfl2025flux2klein}, each reference $k$ receives a dedicated time-axis offset $\delta_k=k\cdot s_{\mathrm{scale}}$, so that attention similarity between references reflects content rather than incidental positional proximity.

\noindent\textbf{Phase 1: isolated grounding ($0\to T/2$).} For reference $k$, we run a forward pass that contains only that reference's tokens,
\begin{equation}
\hat{\tau}^{t+1}_{I_0,k} = \mathcal{A}\!\left( \tau^{t}_{I_0};\; \tau_{P_k},\; \tau_{R_k},\; \bar{\tau}^{\setminus k}_{I_0} \right),
\label{eq:phase1}
\end{equation}
\claude{where $\mathcal{A}$ is one sampling step of the MM-DiT backbone (a transformer forward pass over the concatenated sequence followed by the flow-matching~\cite{lipman2023flowmatching} Euler update), arguments after the semicolon are conditioning tokens, and $\bar{\tau}^{\setminus k}_{I_0}=\mathcal{E}_i\bigl(I_0\otimes(1-M_k)\bigr)$ is the layout with subject $k$'s region blanked, which gives the pass the context around the region it fills. Phase~1 uses no attention mask; $M_k$ acts only through this context and the blend below.} With no other reference in the sequence, \claude{the prediction for the subject's region is conditioned on that subject alone.} The $K$ predictions are composed via mask-weighted spatial blending:
\begin{equation}
\tau^{t+1}_{I_0} = \tilde{\tau}^{t+1}_{I_0} \otimes \Bigl(1 - \bigoplus_{k=1}^{K} M_k\Bigr) \;+\; \sum_{k=1}^{K} \hat{M}_k \otimes \hat{\tau}^{t+1}_{I_0,k},
\label{eq:blend}
\end{equation}
where $\otimes$ is elementwise multiplication, $\bigoplus$ the elementwise maximum, and $\tilde{\tau}^{t+1}_{I_0}$ is $\bar{\tau}_{I_0}$ re-noised to the noise level of step $t{+}1$. \claude{Overlaps go to the first mask in a fixed order, $\hat{M}_k=M_k\otimes\prod_{k'<k}(1-M_{k'})$: subjects (face masks) first, then worn clothing, then all other objects, with ties inside a group broken by Stage~1 box order. The weights therefore form a partition of unity; as in blended diffusion~\cite{Avrahami_2022_CVPR}, the spliced canvas stays in-distribution and the background is re-projected onto the verified layout rather than drifting.}

\noindent\textbf{Phase 2: routed joint refinement ($T/2\to T$).} \claude{The hard assignment of Phase~1 leaves a distribution shift at the seams between independently predicted patches. Phase~2 repairs it with a single joint pass over all references under the global prompt,}
\begin{equation}
  \tau^{t+1}_{I_0} = \mathcal{A}\!\left( \tau^{t}_{I_0};\; \tau_{P_0},\; \tau_{R_1},\, \ldots,\, \tau_{R_K},\; M^{*} \right),
  \label{eq:phase2}
\end{equation}
\claude{which harmonizes lighting, pose, and contact between subjects. It relies on a correspondence that Phase~1 establishes: once each subject's features occupy their region, joint attention weights the matching reference most, so the pass sharpens identity region by region instead of mixing references.}

\begin{table*}[t]
  \centering
  \caption{\claude{\textbf{Per-group-size results on \bench.} Mean over the prompts of each group size, one fixed seed per prompt. \revise{\tablelegend{} All eight methods are ranked.} ``1-pass'' and SRI are training-free controls on our backbone (\cref{sec:exp:setup}). BG-Sim is undefined (--) for methods without a background-image input.}}
  \label{tab:main}
  \vspace{-2mm}
  \footnotesize
  \setlength{\tabcolsep}{1.7pt}
  \begin{minipage}[t]{0.495\textwidth}
  \centering
  \begin{tabular}{@{}c l cccccc@{}}
    \toprule
    \textbf{$N$} & \textbf{Method} & \textbf{ID}$\uparrow$ & \textbf{DINO}$\uparrow$ & \textbf{CLIP-T}$\uparrow$ & \textbf{BG}$\uparrow$ & \textbf{Aesth.}$\uparrow$ & \textbf{Attr-VQA}$\uparrow$ \\
    \midrule
    \multirow{8}{*}{2}
     & OmniGen2 & 13.07 & \third{33.16} & \third{62.23} & -- & 51.63 & 33.04 \\
     & UNO & 23.23 & 28.03 & 62.11 & -- & 55.24 & 35.12 \\
     & DreamO & 3.07 & 5.98 & 58.79 & -- & 44.42 & 14.45 \\
     & XVerse & \third{42.33} & 32.96 & 61.60 & -- & 39.73 & 7.80 \\
     & MOSAIC & 9.93 & 4.77 & 57.29 & -- & 44.01 & 11.22 \\
    \cmidrule(l){2-8}
     & 1-pass & 37.56 & \second{33.56} & 59.88 & \second{0.798} & \best{61.04} & \third{46.57} \\
     & SRI & \second{45.28} & 32.02 & \second{64.07} & \third{0.732} & \third{55.33} & \second{53.41} \\
     & \oursc{\textbf{\method}} & \best{50.58} & \best{34.02} & \best{69.07} & \best{0.873} & \second{58.33} & \best{63.07} \\
    \midrule
    \multirow{8}{*}{3}
     & OmniGen2 & 0.62 & 1.74 & 57.57 & -- & 37.11 & 3.68 \\
     & UNO & 19.60 & 17.35 & \third{61.02} & -- & \third{56.01} & 14.65 \\
     & DreamO & 0.39 & 1.20 & 57.47 & -- & 41.48 & 2.99 \\
     & XVerse & \third{27.21} & 15.62 & 60.08 & -- & 38.46 & 4.03 \\
     & MOSAIC & 11.45 & 5.87 & 57.52 & -- & 42.66 & 5.67 \\
    \cmidrule(l){2-8}
     & 1-pass & 23.18 & \third{29.27} & 60.06 & \second{0.692} & \second{58.98} & \second{41.40} \\
     & SRI & \second{32.61} & \second{30.14} & \second{61.23} & \third{0.633} & 55.21 & \third{37.18} \\
     & \oursc{\textbf{\method}} & \best{54.90} & \best{34.84} & \best{67.89} & \best{0.854} & \best{67.27} & \best{56.28} \\
    \bottomrule
  \end{tabular}
  \end{minipage}\hfill
  \begin{minipage}[t]{0.495\textwidth}
  \centering
  \begin{tabular}{@{}c l cccccc@{}}
    \toprule
    \textbf{$N$} & \textbf{Method} & \textbf{ID}$\uparrow$ & \textbf{DINO}$\uparrow$ & \textbf{CLIP-T}$\uparrow$ & \textbf{BG}$\uparrow$ & \textbf{Aesth.}$\uparrow$ & \textbf{Attr-VQA}$\uparrow$ \\
    \midrule
    \multirow{8}{*}{4}
     & OmniGen2 & 0.39 & 1.60 & 57.63 & -- & 37.36 & 5.78 \\
     & UNO & \third{14.37} & \third{13.90} & \best{60.79} & -- & 50.08 & \third{14.78} \\
     & DreamO & 0.19 & 0.75 & 57.39 & -- & 42.34 & 4.63 \\
     & XVerse & 2.67 & 2.63 & 57.71 & -- & 36.54 & 3.17 \\
     & MOSAIC & 2.46 & 1.99 & 57.34 & -- & 37.73 & 2.61 \\
    \cmidrule(l){2-8}
     & 1-pass & 10.91 & 9.64 & 58.14 & \second{0.672} & \third{51.38} & 8.54 \\
     & SRI & \second{20.14} & \second{19.48} & \third{58.32} & \third{0.611} & \second{56.12} & \second{35.52} \\
     & \oursc{\textbf{\method}} & \best{26.80} & \best{29.28} & \second{60.13} & \best{0.744} & \best{64.18} & \best{44.25} \\
    \midrule
    \multirow{8}{*}{5}
     & OmniGen2 & 0.08 & 0.43 & 57.38 & -- & 36.85 & 2.49 \\
     & UNO & \third{13.47} & \third{8.58} & \second{59.26} & -- & \second{56.40} & \third{7.27} \\
     & DreamO & 0.03 & 0.17 & 57.24 & -- & 37.35 & 1.79 \\
     & XVerse & 0.52 & 0.46 & 57.15 & -- & 36.66 & 1.28 \\
     & MOSAIC & 0.18 & 0.24 & 57.24 & -- & 36.79 & 1.92 \\
    \cmidrule(l){2-8}
     & 1-pass & 1.58 & 2.32 & 57.79 & \third{0.401} & 42.38 & 3.43 \\
     & SRI & \second{18.31} & \second{16.72} & \third{57.98} & \second{0.610} & \third{55.21} & \second{9.48} \\
     & \oursc{\textbf{\method}} & \best{25.26} & \best{27.24} & \best{60.98} & \best{0.807} & \best{63.81} & \best{40.27} \\
    \bottomrule
  \end{tabular}
  \end{minipage}
  \vspace{-3mm}
\end{table*}

\noindent\textbf{Deterministic routing mask.} \claude{Residual shift near mask boundaries can still leave output tokens attending to several references, so the joint pass carries a routing mask $M^{*}$ (visualized in Supp.~B.5).} 
\claude{Given the sequence $[\,\text{text}\mid\text{output}\mid R_1\mid\cdots\mid R_K\mid R_{\mathrm{bg}}\,]$, where $R_{\mathrm{bg}}$ is the background reference,} let $\mathcal{O}$ and $\mathcal{R}_k$ be the index sets of the output tokens and of reference $k$. Each output token $i$ gets an assignment $a(i)\in\{0,1,\dots,K\}$: \claude{$a(i)=k$ if $M_k$ covers more than half of its token-grid cell (the mask with the largest coverage wins where masks overlap), and $a(i)=0$ for background tokens that no mask covers.} \claude{The additive mask, broadcast over heads, is}
\begin{equation}
  M^{*}_{i,j} =
  \begin{cases}
    -\infty & i\in\mathcal{O},\ a(i)\neq 0,\ j\in\mathcal{R}_k\cup\mathcal{R}_{\mathrm{bg}},\ k\neq a(i),\\
    0       & \text{otherwise,}
  \end{cases}
  \label{eq:mask}
\end{equation}
\claude{so a subject's output token attends only to its own reference among all references, including the background reference. Every other entry stays open: background output tokens ($a(i)=0$) attend to all references, and reference rows, text tokens, and output-to-output attention are unrestricted. $M^{*}$ is applied in every double-stream block of the Phase-2 joint pass, and the single-stream blocks keep full attention. \revise{This placement follows from the role of Phase~2: Phase~1 has already grounded each subject, so Phase~2 only needs to stop a region from reading the wrong reference while the scene is harmonized. Reference rows and the single-stream blocks, where text and image tokens share one stream, stay open to keep the global context needed for lighting, pose, and contact.}}

\claude{\noindent\textbf{Scope of the guarantee.} \cref{eq:mask} stops every subject region from reading another subject's reference directly, but reference rows, output-to-output and text attention, and the single-stream blocks remain open, so information can still propagate indirectly across layers. We keep them open on purpose, since hard boundaries would also block legitimate exchange across contact regions (a hand on a shoulder), and \revise{Supp.~B.6 visualizes where output-to-reference attention falls with and without $M^{*}$}. Unlike online winner-take-all routing~\cite{Hsiao2025TFTI2ITT}, the assignment is fixed before sampling, so it cannot oscillate between similar subjects.}

\section{Experiments}
\label{sec:exp}

\subsection{Setup}
\label{sec:exp:setup}


\noindent\textbf{Methods.} We evaluate five published multi-subject generators, OmniGen2~\cite{omnigen2}, UNO~\cite{wu2025uno}, DreamO~\cite{dreamo2025}, XVerse~\cite{xverse2025}, and MOSAIC~\cite{mosaic2025}, with their official checkpoints and default sampling settings. Two training-free controls share our FLUX.2-klein-9B~\cite{bfl2025flux2klein} backbone: a \emph{single-pass} control that concatenates all reference tokens into one call, and \emph{SRI}, \revise{a sequential-inpainting control that we construct and release with \bench. SRI inpaints subjects one at a time into an anonymous-placeholder layout, so each pass sees a single reference (Supp.~A.6). It therefore tests per-subject decomposition alone, whereas \method keeps the decomposition and adds scene-graph verification of the layout and joint two-phase rendering. Because SRI is our own construction, we report it next to the published methods but exclude it when we compute margins over the strongest published baseline.}


\noindent\textbf{Implementation.} Object Parsing, Scene Graph, Prompt Enhancement, and Prompt Refinement agents share one Qwen2.5-VL-7B~\cite{bai2025qwen25vl} instance under different system prompts. Masks come from Rex-Omni~\cite{jiang2026rexomni} detection and SAM ViT-H~\cite{kirillov2023sam} segmentation, with face regions parsed by BiSeNet~\cite{yu2018bisenet}. Sampling uses $T{=}16$ flow-matching steps with guidance scale 1.5 and the phase split at step~8\revise{, the midpoint of the schedule}; the layout loop uses $\theta{=}0.8$ and $T_{\max}{=}3$. The Attr-VQA judge (\texttt{gemma-4-e4b-it}) shares no model with the generation pipeline. Full configurations are in Supp.~C.

\subsection{Benchmark Results}
\label{sec:exp:main}

\begin{table}[t]
  \centering
  \caption{\claude{\textbf{Averages over group sizes}, weighted by prompt share (41.0/28.1/16.5/14.4\%), computed from \cref{tab:main}. \revise{Shading as in \cref{tab:main}.}}}
  \label{tab:avg}
  \vspace{-2mm}
  \footnotesize
  \setlength{\tabcolsep}{2.0pt}
  \claude{%
  \begin{tabular}{@{}l cccccc@{}}
    \toprule
    \textbf{Method} & \textbf{ID}$\uparrow$ & \textbf{DINO}$\uparrow$ & \textbf{CLIP-T}$\uparrow$ & \textbf{BG}$\uparrow$ & \textbf{Aesth.}$\uparrow$ & \textbf{Attr-VQA}$\uparrow$ \\
    \midrule
    OmniGen2~\cite{omnigen2} & 5.61 & 14.41 & 59.46 & -- & 43.07 & 15.89 \\
    UNO~\cite{wu2025uno} & 19.35 & 19.91 & \third{61.17} & -- & 54.77 & 22.00 \\
    DreamO~\cite{dreamo2025} & 1.40 & 2.94 & 57.96 & -- & 42.23 & 7.79 \\
    XVerse~\cite{xverse2025} & \third{25.51} & 18.40 & 59.89 & -- & 38.40 & 5.04 \\
    MOSAIC~\cite{mosaic2025} & 7.72 & 3.97 & 57.36 & -- & 41.55 & 6.90 \\
    \midrule
    1-pass (FLUX.2) & 23.94 & \third{23.90} & 59.34 & \third{0.649} & \second{56.18} & \third{32.62} \\
    SRI & \second{33.68} & \second{27.22} & \second{61.45} & \second{0.667} & \third{55.41} & \second{39.57} \\
    \oursc{\textbf{\method}} & \best{44.22} & \best{32.49} & \best{66.10} & \best{0.819} & \best{62.60} & \best{54.77} \\
    \bottomrule
  \end{tabular}}
  \vspace{-3mm}
\end{table}

\claude{\cref{tab:main,tab:avg} report per-$N$ and averaged results. We read them first as a benchmark, then as an evaluation of \method.}


\noindent\textbf{Published generators collapse with group size.}
\revise{XVerse, the strongest published method on identity at $N{=}2$, drops from 42.33 to 0.52 ID-Sim at $N{=}5$ (81-fold), and OmniGen2 and DreamO reach near-zero ID-Sim for $N\ge3$. At $N\ge4$, four of the five published methods score at most 2.63 DINO-Sim, and between two and five subjects every published method loses 79--92\% of its Attr-VQA.}

\begin{figure*}[t]
  \centering
  \includegraphics[width=0.9\textwidth]{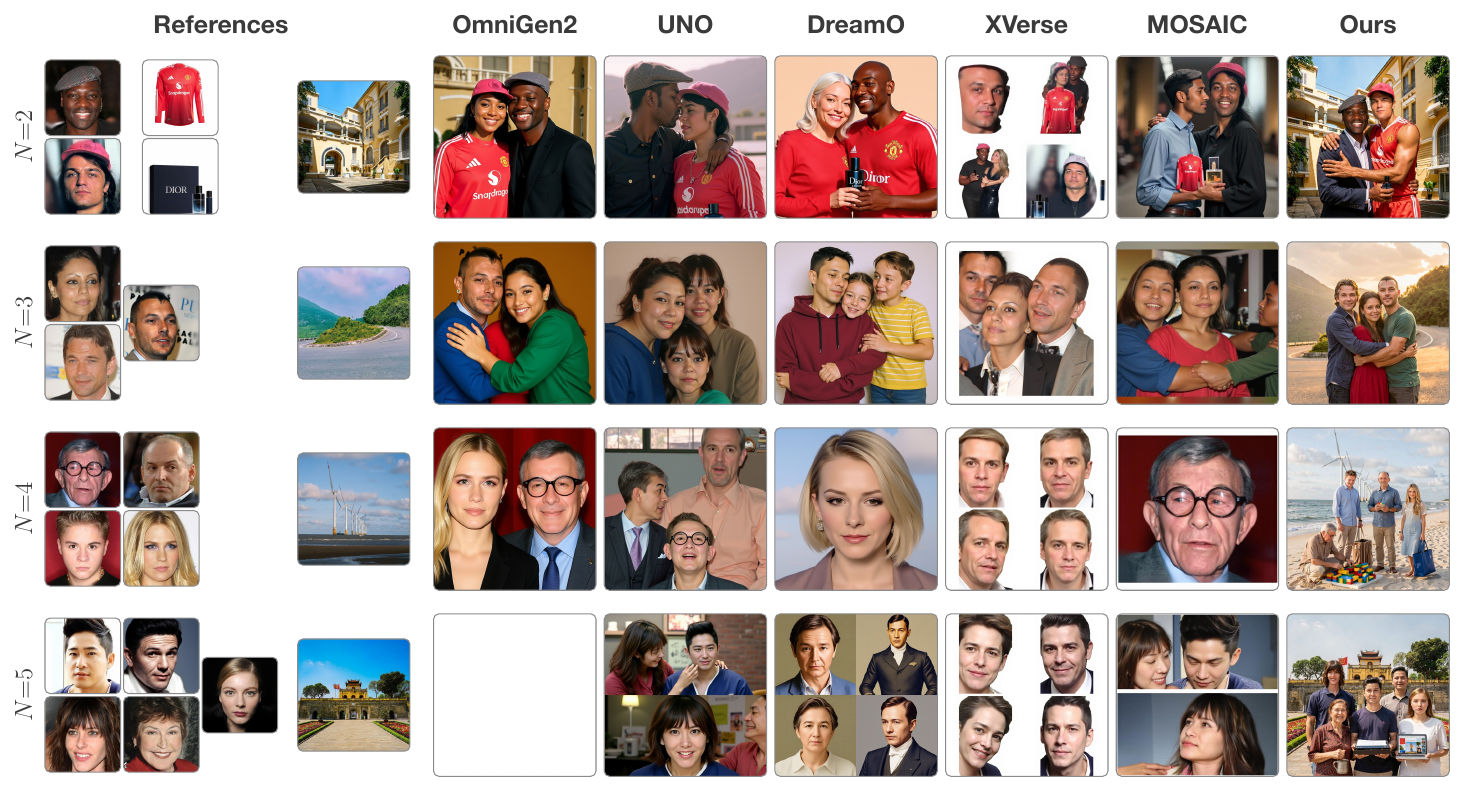}
  \vspace{-3mm}
  \caption{\textbf{Qualitative comparison on \bench.} One scene per group size (rows); every method (columns) receives the identical prompt and subject references\claude{, and the same background (as an image for ours, as text for the published baselines)}, grouped at left. OmniGen2 returns a blank canvas for every $N{=}5$ prompt, hence its empty tile.}
  \label{fig:qualitative}
  \vspace{-3mm}
\end{figure*}

\noindent\textbf{Global metrics hide the collapse.}
\revise{CLIP-T stays between 57.15 and 62.23 across all published methods and group sizes. UNO's aesthetic score is similar at two and five subjects (55.24 vs.\ 56.40) while its Attr-VQA falls from 35.12 to 7.27, and XVerse reaches 42.33 ID-Sim but only 7.80 Attr-VQA at $N{=}2$ by rendering disconnected portraits (\cref{fig:qualitative}). Global scores therefore need per-subject identity and binding measures beside them.}

\claude{\noindent\textbf{Decomposition helps but is not enough.}} \revise{On the same backbone, conditioning each pass on a single subject already helps: SRI raises the single-pass control's average ID-Sim from 23.94 to 33.68 and Attr-VQA from 32.62 to 39.57 (\cref{tab:avg}), and it is the strongest method other than ours. Decomposition alone does not scale, however: SRI's Attr-VQA falls from 35.52 at $N{=}4$ to 9.48 at $N{=}5$, consistent with errors accumulating over $N$ sequential passes, none of which sees the subjects already inserted.}

\noindent\textbf{\method.}
\revise{Our method ranks first on every axis at every group size except CLIP-T at $N{=}4$ (UNO, by 0.66) and Aesth.\ at $N{=}2$ (the single-pass control). On prompt-weighted averages it reaches $2.5\times$ the Attr-VQA of the strongest published baseline (54.77 vs.\ 22.00), $1.7\times$ its ID-Sim, and $1.6\times$ its DINO-Sim (best baseline chosen per metric), against a CLIP-T gain of 4.93 points. Against SRI, which shares our backbone and placeholder layout but neither verifies the layout nor renders the subjects jointly, ID-Sim rises from 33.68 to 44.22 and Attr-VQA from 39.57 to 54.77, with a gap of 30.79 points at $N{=}5$; \cref{tab:ablation} separates the two contributions. Our five-subject Attr-VQA (40.27) still exceeds every published method's two-subject score, and the ID-Sim margin over the single-pass control grows from 13.02 at $N{=}2$ to 23.68 at $N{=}5$, so the gain comes from the pipeline rather than the backbone.}

\subsection{Is Attr-VQA Trustworthy?}
\label{sec:exp:vqa}

\claude{Seven annotators with computer-vision backgrounds answered Yes/No/Unsure to the Attr-VQA questions of 46 generated images (515 image--question pairs, 2,869 valid votes), and the majority vote served as ground truth. At the image level, the judge's Yes rate correlates with the human one (Spearman $\rho=0.645$, Pearson $r=0.656$, Kendall $\tau_b=0.506$); at the question level, it agrees with the human majority on 76.5\% of pairs (Cohen's $\kappa=0.531$).} \claude{It is thus reliable for comparing methods, but optimistic in absolute terms (mean Yes rate 57.64\% vs.\ 51.00\% for humans).}

\subsection{Ablation Study}
\label{sec:exp:ablation}

\begin{table}[t]
  \centering
  \caption{\textbf{Component ablation on \bench.} Each variant removes or replaces one component with all else fixed\claude{, averaged over $N$ with prompt-share weights.} ``w/o verification'' takes the first generated layout, removing both the feedback and up to $T_{\max}{-}1$ extra attempts; the attempt-matched row draws $T_{\max}$ unverified layouts and keeps one without consulting the verifier. \revise{Shading as in \cref{tab:main}.}}
  \label{tab:ablation}
  \vspace{-2mm}
  \footnotesize
  \setlength{\tabcolsep}{2.3pt}
  \resizebox{\linewidth}{!}{
  \begin{tabular}{@{}l cccccc@{}}
    \toprule
    \textbf{Variant} & \textbf{ID}$\uparrow$ & \textbf{DINO}$\uparrow$ & \textbf{CLIP-T}$\uparrow$ & \textbf{BG}$\uparrow$ & \textbf{Aesth.}$\uparrow$ & \textbf{Attr-VQA}$\uparrow$ \\
    \midrule
    \oursc{\textbf{Context2Canvas (Full)}} & \best{44.22} & \oursc{32.49} & \best{66.10} & \third{0.819} & \third{62.60} & \second{54.77} \\
    w/o verification & \third{37.08} & \best{43.39} & 62.74 & 0.759 & 60.86 & 44.78 \\
    \quad attempt-matched & 36.87 & \third{32.58} & 62.70 & 0.757 & 61.63 & 38.37 \\
    w/o Phase~1 & 28.42 & 21.40 & 63.47 & 0.817 & 37.87 & 45.19 \\
    w/o Phase~2 & 29.42 & 14.83 & 37.26 & \best{0.832} & 36.83 & 42.93 \\
    w/o RoPE offset $\delta_k$ & 12.70 & 31.33 & \third{63.93} & 0.815 & \best{63.59} & \third{52.99} \\
    $M^{*}\!\to$ online WTA~\cite{Hsiao2025TFTI2ITT} & \second{38.20} & \second{39.53} & \second{64.26} & \second{0.831} & \second{63.08} & \best{56.72} \\
    \bottomrule
  \end{tabular}
  }
  \vspace{-3mm}
\end{table}


\noindent\textbf{Verification.}
Removing verification reduces ID-Sim from 44.22 to 37.08 and Attr-VQA from 54.77 to 44.78, while DINO-Sim rises to 43.39. Using the same layout-attempt budget without verifier feedback (the attempt-matched control) yields only 36.87 ID-Sim and 38.37 Attr-VQA, supporting verification for identity and binding.


\noindent\textbf{Two phases.}
\revise{Removing Phase~1 lowers ID-Sim from 44.22 to 28.42 and Aesth.\ from 62.60 to 37.87. Removing Phase~2 keeps ID-Sim low (29.42) and drops CLIP-T to 37.26 and Aesth.\ to 36.83. Attr-VQA falls without either phase (45.19 and 42.93 vs.\ 54.77), so isolated grounding and joint refinement are both needed.}

\noindent\textbf{Routing rule.}
\revise{Replacing $M^{*}$ with online winner-take-all routing gives a trade-off: $M^{*}$ is higher on ID-Sim (44.22 vs.\ 38.20) and CLIP-T (66.10 vs.\ 64.26), and WTA on DINO-Sim (39.53 vs.\ 32.49), Aesth.\ (63.08 vs.\ 62.60), and Attr-VQA (56.72 vs.\ 54.77). Deterministic routing thus favors identity and text alignment, without an advantage in binding.}


\noindent\textbf{Cost.}
\revise{The pipeline takes 430.27\,s per image on one A100 with sequential offloading, $5.46\times$ a single FLUX pass (78.81\,s). Layout generation and verification take 78.8\% of this; two-phase rendering adds only ${\sim}16\%$ over a single pass (Supp.~D.2).}


\subsection{Qualitative Results}
\label{sec:exp:qual}

\revise{The published methods in \cref{fig:qualitative} fail in the three patterns that \bench's axes separate. \emph{Partial identity preservation}: UNO keeps a well-composed group whose faces drift. \emph{Attribute leakage}: items are dropped, swapped, or replaced by look-alikes; XVerse tiles the references into separate portraits and MOSAIC blends subjects into fragments. \emph{Background regression}: OmniGen2 and DreamO replace the landmark with a studio backdrop. Only ours returns one coherent scene with every subject, object, and landmark at every $N$, though not perfectly: the long-sleeved jersey at $N{=}2$ becomes sleeveless, and at $N{=}3,4$ the landmark is re-rendered from $I_{bg}$ rather than copied (BG-Sim 0.734).}

\claude{\noindent\textbf{Failure cases.}} \revise{As $N$ grows, identity grounding degrades first: in a verified five-subject layout, a reference can bind to the wrong subject or yield a featureless face. As verification checks presence only (\cref{sec:method:layout}), a layout with implausible subject scale still passes (Supp.~F.2).}

\subsection{Human Preference Study}
\label{sec:exp:human}

\revise{Twelve raters who passed attention checks made 972 judgments over 324 blind pairwise comparisons against UNO, OmniGen2, XVerse, and MOSAIC. They prefer ours on every dimension against every baseline, overall from 77\% (OmniGen2) to 93\% (XVerse), and every head-to-head is significant (two-sided sign test, $p<0.01$). Agreement is low in absolute terms (Krippendorff's~\cite{krippendorff2018content} $\alpha\approx0.30$), as expected for forced subjective choices, but a Bradley--Terry~\cite{bradley1952rank} fit gives confidence intervals disjoint from every baseline (Supp.~E).}


\subsection{\texorpdfstring{\revise{Limitations}}{Limitations}}
\label{sec:exp:limitations}


\revise{CelebA-HQ has limited demographic breadth, and identities and web-sourced items may overlap generator training data, so absolute ID-Sim and DINO-Sim should be read comparatively. The graphs and the Attr-VQA judge share the Gemma~4 family. Verification checks entity presence rather than relations (\cref{sec:exp:qual}) and inherits detector and segmenter errors, and routing assumes masks that cover each subject. The added latency confines \method to offline use.}

\section{\texorpdfstring{\revise{Conclusion}}{Conclusion}}
\label{sec:conclusion}

\revise{\bench makes every identity, binding, position, interaction, and background requirement separately checkable, and shows that published generators collapse as subjects are added while global metrics stay flat. \method turns a verified layout into a deterministic routing mask for a two-phase renderer, improving attribute binding by $2.5\times$ and identity by $1.7\times$ over the strongest published method. Verifying relations and reducing cost are the next steps.}


\section*{LLM Usage Statement}

\revise{LLMs enter this work in two distinct roles, and we report both.

\noindent\textbf{As research components.} Some LLMs and MLLMs are part of the benchmark and the method, and their outputs are part of the reported data and results by design. Gemma~4 (\texttt{gemma-4-e4b-it}) extracts reference attributes, synthesizes and verifies the \bench prompts and their interaction and position graphs (\cref{sec:bench:construction}), and serves as the Attr-VQA judge (\cref{sec:bench:metrics}). Its agreement with human majority judgments is measured in \cref{sec:exp:vqa}, and because the same model family also wrote the prompts, we release the full prompt and answer logs so that the benchmark can be re-scored with other judges. Qwen2.5-VL-7B runs the Object Parsing, Scene Graph, Prompt Enhancement, and Prompt Refinement agents of \method (\cref{sec:exp:setup}). No model in the generation pipeline is used as a judge.

\noindent\textbf{As writing and coding assistants.} Claude (Anthropic) assisted with debugging code, with editing the language of the manuscript and supplement, including restructuring and rewording some paragraphs, and with checking consistency between the text, tables, and figures. The authors reviewed every suggestion, revised it where needed, and decided whether to adopt it. No assistant ran experiments, produced or altered any reported number, or proposed the research ideas, the method, or the experimental design.

The authors are responsible for the data, analyses, and text, and take full responsibility for the accuracy, originality, and integrity of the paper.}

{
    \small
    \bibliographystyle{ieeenat_fullname}
    \bibliography{cvpr_ref}
}

\clearpage
\maketitlesupplementary

\appendix

\setcounter{figure}{0}
\setcounter{table}{0}
\setcounter{equation}{0}
\setcounter{algorithm}{0}
\renewcommand{\thefigure}{S\arabic{figure}}
\renewcommand{\thetable}{S\arabic{table}}
\renewcommand{\theequation}{S\arabic{equation}}
\renewcommand{\thealgorithm}{S\arabic{algorithm}}
\renewcommand{\theHfigure}{S\arabic{figure}}
\renewcommand{\theHtable}{S\arabic{table}}
\renewcommand{\theHequation}{S\arabic{equation}}
\providecommand{\theHalgorithm}{}
\renewcommand{\theHalgorithm}{S\arabic{algorithm}}

\claude{\noindent This supplement provides the material deferred from the main paper. \cref{sec:supp:bench} details the \bench benchmark: curation, prompt synthesis, statistics, metric implementation, the Attr-VQA human study, and the SRI baseline. \cref{sec:supp:method} details \method: the retrieval funnel, mask preprocessing, the verification loop, the routing mask, attention diagnostics, and the agent inventory. \cref{sec:supp:impl} gives the configuration and compute behind every reported number. \cref{sec:supp:quant,sec:supp:human} add quantitative results and the human preference protocol, \cref{sec:supp:qual} adds qualitative results, and \cref{sec:supp:impact} discusses \revise{broader impact, ethics,} and release.}

\section{\bench Benchmark Details}
\label{sec:supp:bench}

\subsection{Reference Curation}
\label{sec:supp:curation}

Person identities are drawn from CelebA-HQ~\cite{karras2018progressive} and balanced along gender, age group, and ethnicity by a two-annotator audit. Object and fashion candidates are deduplicated with DINOv2~\cite{oquab2023dinov2} on \texttt{[CLS]} features (cosine threshold $0.90$), filtered by the LAION aesthetic predictor v2.5~\cite{schuhmann2022laion} at threshold $5.0$, and screened by a CLIP ViT-L/14~\cite{radford2021clip} text--image alignment check against the named query at threshold $0.25$. A final manual review enforces the distinctiveness and image-quality criteria from \claude{Sec.~3.2}. The 29 background scenes are individually reviewed for global recognizability, sufficient resolution ($\geq\!1024$\,px on the short side), and absence of incidental persons that would conflict with subject insertion. \claude{Distinctive interiors would satisfy the identifiability criterion, but generic rooms are outside the current pool.}

\subsection{Prompt Synthesis and Verification}
\label{sec:supp:prompt}

Both attribute extraction and prompt synthesis use Gemma\,4-e4b-it~\cite{gemma2026gemma4}, decoded with \texttt{temperature}=0 under a JSON-schema-constrained generator. For every reference asset Gemma\,4 emits structured fields---\texttt{age}, \texttt{gender}, \texttt{ethnicity} for identities; \texttt{size\_category} and \texttt{afforded\_interactions} for objects/fashion; \texttt{scene\_type} and \texttt{landmark\_name} for backgrounds---which condition the compositional sampler. Each candidate prompt is then passed through three Gemma\,4 verification queries (interaction plausibility, layout feasibility, scene compatibility); the per-stage rejection rates are 9.4\%, 6.1\%, and 3.7\%, and 1.2\% of prompts require a human override. The final yield is 1{,}361 prompts \claude{from about 4{,}000 candidates}. \claude{Positions are assigned by sampling an ordering of the $N$ identities over the layout slots, so that across the prompt set every identity appears in every slot rather than in a fixed canonical order.} \claude{Human review takes place during dataset construction only, never during model inference.}

\subsection{Statistics and Benchmark Comparison}
\label{sec:supp:stats}

\begin{figure*}[t]
  \centering
  \includegraphics[trim={40mm 22mm 40mm 0},clip, width=\textwidth]{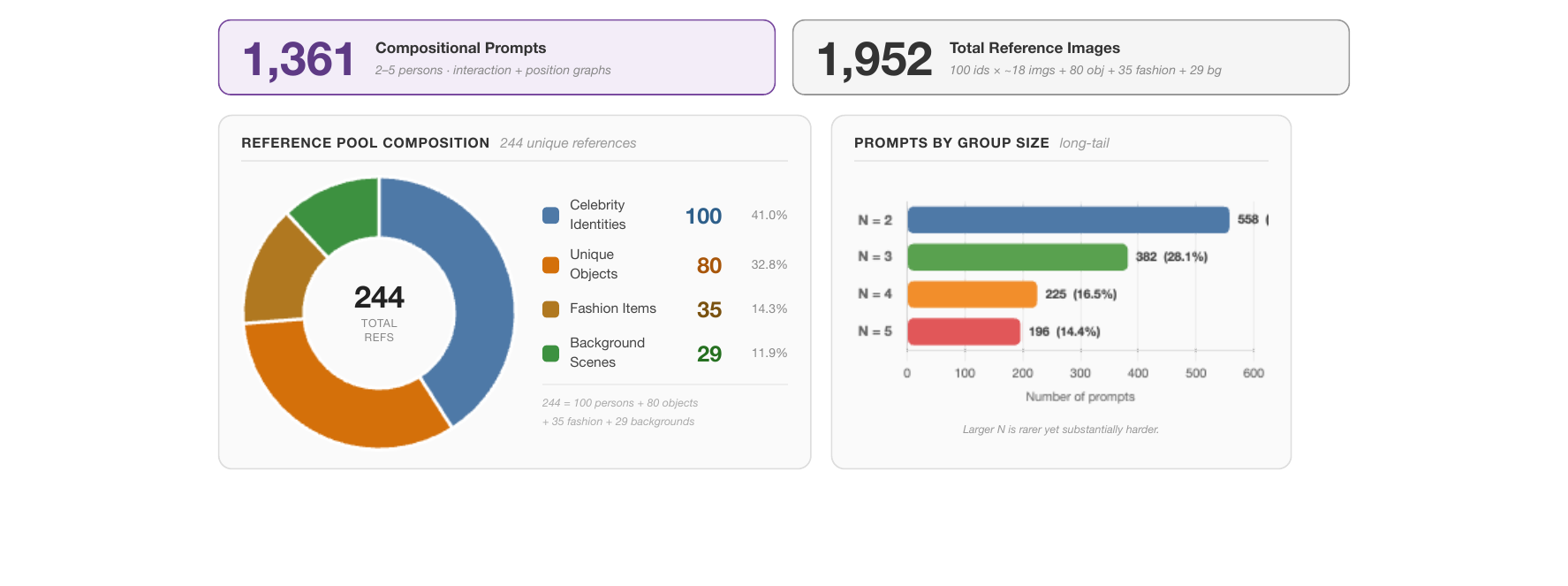}
  \vspace{-5mm}
  \caption{\textbf{\bench dataset statistics.}
  \emph{Top:} Total compositional prompts (1,361) and reference images (1,952).
  \emph{Left:} Composition of the 244 unique references by category: 100 celebrity identities (41.0\%), 80 unique objects (32.8\%), 35 fashion items (14.3\%), and 29 background scenes (11.9\%).
  \claude{\emph{Right:}} Long-tail prompt distribution across group sizes $N=2$--5.}
  \label{fig:supp:stats}
  \vspace{-3mm}
\end{figure*}

\begin{table}[t]
  \centering
  \small
  \caption{\bench composition. The group-size shares are the prompt-share weights used for every average over $N$ \claude{in the main paper (Tabs.~3 and~4)}.}
  \label{tab:supp:stats}
  \begin{tabular}{@{}lr@{}}
    \toprule
    \textbf{Statistic} & \textbf{Count} \\
    \midrule
    Celebrity identities & 100 \\
    \quad \claude{reference images per identity} & \claude{${\approx}18$} \\
    Unique objects & 80 \\
    Fashion items & 35 \\
    Background scenes & 29 \\
    Reference images & 1{,}952 \\
    Compositional prompts & 1{,}361 \\
    \midrule
    \quad 2-person prompts & 558 (41.0\%) \\
    \quad 3-person prompts & 382 (28.1\%) \\
    \quad 4-person prompts & 225 (16.5\%) \\
    \quad 5-person prompts & 196 (14.4\%) \\
    \bottomrule
  \end{tabular}
\end{table}

\cref{fig:supp:stats} and \cref{tab:supp:stats} summarize the benchmark. \claude{The 1{,}952 reference images comprise about 18 images for each of the 100 identities plus one image for each of the 80 objects, 35 fashion items, and 29 scenes.} A distinctive property of \bench is that \emph{each identity is paired with a pool of reference images} captured under varied lighting, pose, and apparel, so that different references can be attribute-matched to a given prompt\claude{: a} subject described as ``wearing a red jacket'' should be grounded to the reference that shows that outfit, not an arbitrary one. This design enables the reference retrieval task \revise{described in \cref{sec:supp:metrics}}.

\claude{\noindent\textbf{Subject counts in Tab.~1.} DreamBench = 30 subjects (9 animals + 21 objects) $\times$ 25 prompts; DreamBench++ = 150 subjects (20 humans + 45 animals + 65 objects + 20 styles) $\times$ 9 prompts; XVerseBench = 139 subjects (20 humans + 45 animals + 74 objects) across 300 prompts. The comparison emphasizes joint coverage rather than prompt count: each \bench prompt carries several separately verifiable identity, binding, position, and interaction assertions, so the number of checked relations is much larger than the number of prompts.}

\subsection{Metric Implementation}
\label{sec:supp:metrics}

\claude{The main paper defines the six axes (Sec.~3.3); this section states how each is computed, including the reference-to-instance assignment rule, which is shared by two axes and most affects how the numbers should be read.}

\noindent\textbf{Assignment is best-match, not one-to-one.} For both ID-Sim and DINO-Sim, every reference is scored against \emph{all} detected instances in the generated image and keeps the maximum:
\begin{equation}
  \mathrm{score}(k) = \max_{g \in \mathcal{G}} \, \mathrm{sim}\bigl(I_{ref_k},\, g\bigr),
  \qquad
  \mathrm{score}(k) = 0 \text{ if } \mathcal{G} = \varnothing,
  \label{eq:supp:assign}
\end{equation}
where $\mathcal{G}$ is the set of detected faces or object instances. No one-to-one constraint is imposed, so multiple references may select the same generated subject, each receiving its own similarity score. Applied identically to every method, this permissive rule makes the collapse below one point at $N \geq 3$ evidence of low best-match identity similarity rather than an unfavourable pairing; it does not establish that no usable faces were rendered. \revise{With nonnegative similarities, unmatched references scored as zero, and identical averaging over references, \cref{eq:supp:assign} upper-bounds the corresponding Hungarian score for every method. It does not penalize instance reuse and can therefore mask failures to render distinct subjects.}

\noindent\textbf{ID-Sim.} Faces are detected and tightly warped in both the reference and the generated image, \claude{parsed with facexlib BiSeNet~\cite{yu2018bisenet} to exclude neck and clothing,} embedded with ArcFace~\cite{deng2019arcface} IR-SE50, and compared by cosine similarity scaled to $[0,100]$. Detection failure on either side returns zero.

\noindent\textbf{DINO-Sim.} Object and fashion instances are extracted from the generated image by label using Florence-2~\cite{xiao2024florence2} grounding followed by SAM~\cite{kirillov2023sam} segmentation, with instances below 64\,px discarded, then embedded with DINOv2 and compared against the reference crop. Assignment follows \cref{eq:supp:assign}.

\noindent\textbf{BG-Sim.} SAM~3~\cite{carion2025sam3} masks every foreground subject in both the generated image and the reference background, and DINOv3~\cite{simeoni2025dinov3} cosine similarity is taken over the remaining pixels \claude{and averaged over prompts}. Masking both sides is what prevents a subject rendered over a correct background from inflating the score.

\noindent\textbf{CLIP-T and Aesth.} CLIP-T uses CLIP ViT-L/14 on the prompt and full image; Aesth.\ uses the LAION aesthetic predictor v2.5.

\noindent\textbf{Attr-VQA.} The judge is Gemma\,4-e4b-it, decoded with \texttt{temperature}=0 and \texttt{max\_tokens}=2 against a binary Yes/No template. \claude{One query is issued per subject--attribute entry and per subject-pair relation entry of the prompt's graphs, so a prompt with $N$ subjects costs $O\bigl(N+\binom{N}{2}\bigr)$ queries, and the score is the percentage answered affirmatively. Because every question asserts a ground-truth relation, this equals accuracy against the expected answer.}

\noindent\textbf{Auxiliary tasks.} \claude{Text-only generation is scored by person-count accuracy (person-detector count against the ground-truth $N$) and by Attr-VQA restricted to attribute-binding queries. \revise{Beyond the tasks evaluated in the main paper, the benchmark also supports a reference-retrieval task: because each identity ships with a pool of attribute-varied images, a retriever can be asked to choose the best-matching reference for each prompt slot and scored by Recall@$k$, $k\in\{1,5\}$, against the annotated attribute-matched image.}}

\begin{figure*}[t]
  \centering
  \includegraphics[width=0.9\linewidth]{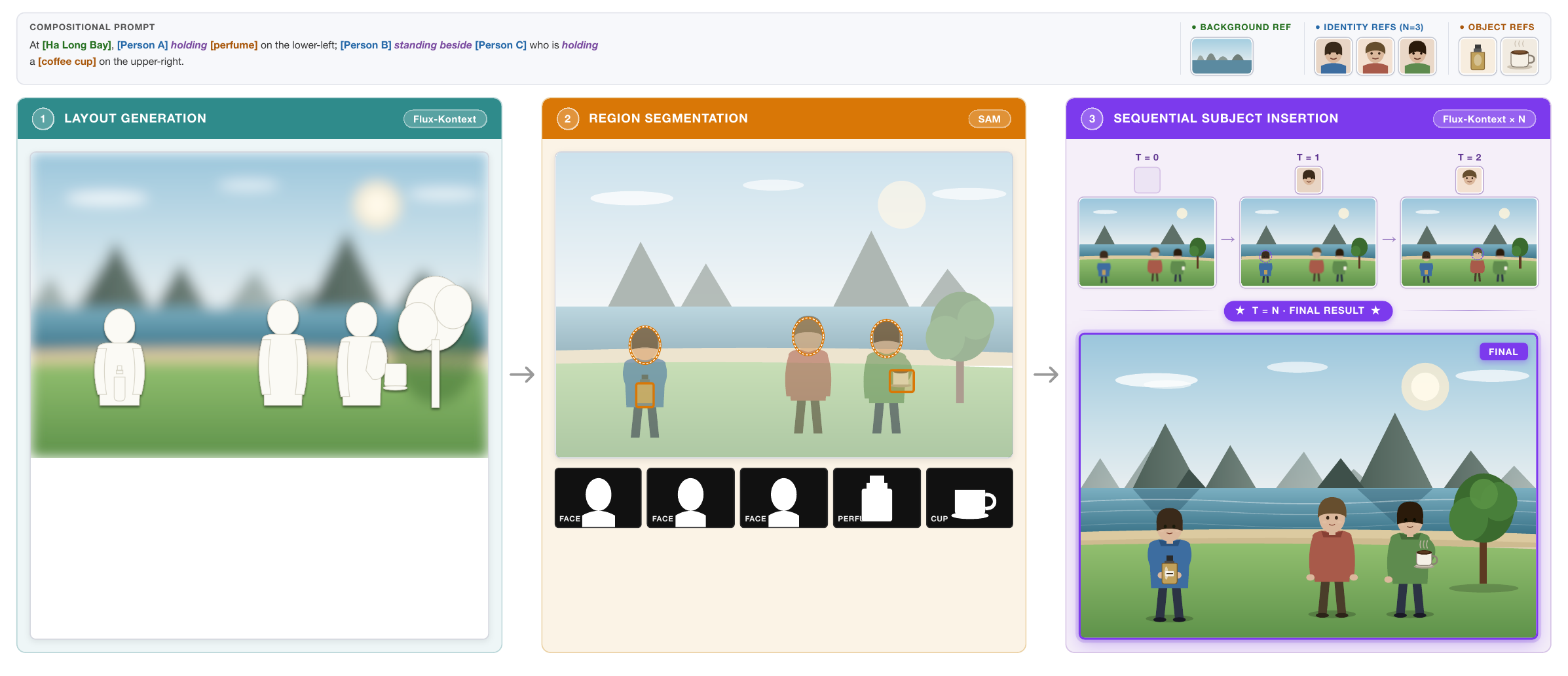}
  \vspace{-3mm}
  \caption{\textbf{SRI (Sequential Reference Inpainting), the training-free baseline released with \bench.} Given a compositional prompt with background, identity, and object/fashion references, FLUX.2-klein generates a scene layout with subject placeholders, SAM segments the face/object regions, and FLUX.2-klein inpaints each region in turn, conditioned on its own reference, to reduce cross-subject feature leakage.}
  \label{fig:supp:sri}
  \vspace{-3mm}
\end{figure*}

\subsection{Human Validation of Attr-VQA}
\label{sec:supp:vqa_human}

\claude{The study of Sec.~5.3 used seven annotators with backgrounds in computer vision and image generation, who independently answered Yes, No, or Unsure to the Attr-VQA questions of 46 generated images. Of the votes on 515 image--question pairs, 2{,}869 remained after removing Unsure responses. The human label of each question is its majority vote. Two caveats apply. Human votes were evaluated against a majority label that includes each rater's own vote, so any human-agreement figure derived from them is not a leave-one-rater-out estimate. And the study measures agreement on images from the methods it sampled, not on every method in the main table.}


\subsection{SRI: Sequential Reference Inpainting Baseline}
\label{sec:supp:sri}

\revise{Sequential Reference Inpainting (SRI) is a training-free control that we construct and release with \bench, so that later work has a reproducible reference point that, unlike the published generators, accepts identity, object/fashion, and background references together (\cref{fig:supp:sri}). It is not a contribution of this paper. Nor is it an ablation of \method: the two share the placeholder idea but differ in several components at once, so SRI answers a narrower question, namely how far per-subject decomposition alone goes on the same backbone.} \emph{Stage~1} conditions FLUX.2-klein~\cite{bfl2025flux2klein} on the background reference and on the compositional prompt with all identity names replaced by anonymous tokens, producing a base image at $1024\!\times\!1024$ with $N$ anonymous placeholders. \emph{Stage~2} segments face and object regions for each subject slot with SAM~\cite{kirillov2023sam}. \emph{Stage~3} inpaints subjects sequentially: for each $i= 1,\dots,N$, FLUX.2-klein runs masked diffusion conditioned on the subject's reference, then the canvas is re-segmented before the next subject. A single retry with seed~$+1$ guards against detection drops; persistent failures contribute zeros to per-subject metrics. \revise{SRI and \method differ in three respects. (i) \method verifies the layout against a scene graph before injecting identities, whereas SRI uses its first layout. (ii) \method replaces the $N$ sequential inpainting passes with the two-phase renderer and routing mask, so that subjects are harmonized jointly rather than pasted one at a time. (iii) The retry budgets differ: SRI retries once on a detection drop, whereas \method draws up to $T_{\max}{=}3$ layouts. The attempt-matched control of Tab.~4 shows that extra layout attempts without verifier feedback do not account for the gain (38.37 Attr-VQA, below SRI's 39.57). The sharp drop of SRI's Attr-VQA from $N{=}4$ to $N{=}5$ (Tab.~2) is consistent with errors accumulating over sequential passes.}

\section{\method Details}
\label{sec:supp:method}

\subsection{Retrieval Funnel}
\label{sec:supp:funnel}

\begin{figure*}[t]
  \centering
  \includegraphics[width=0.9\textwidth]{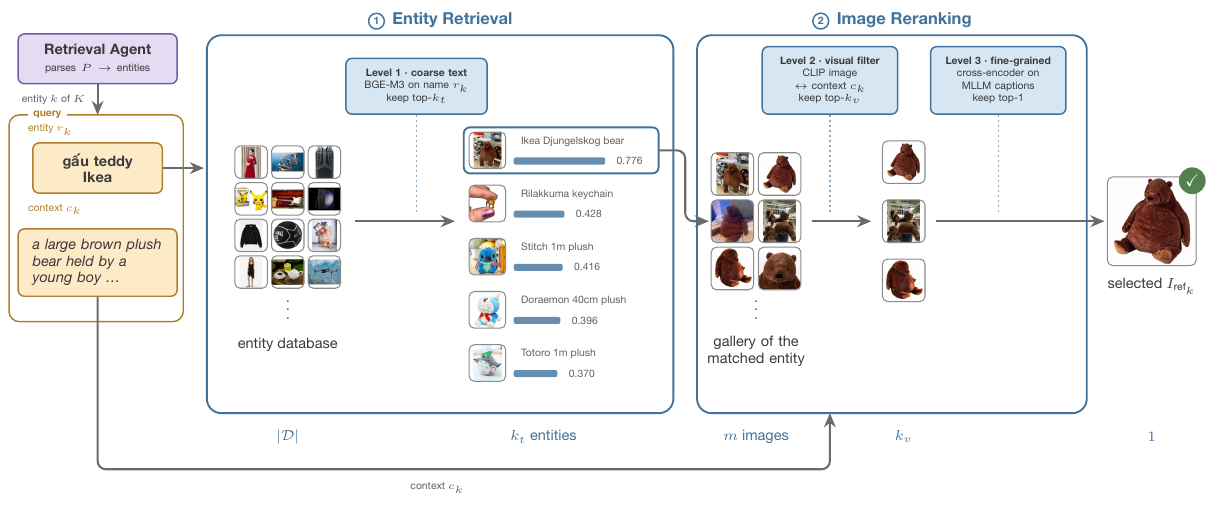}
  \vspace{-5mm}
  \caption{\textbf{The three-level retrieval funnel.} Agent boxes in purple, the parsed query in orange, the two ranked panels in blue. Scores beside each row are retrieval similarities; the counts beneath the arrow give the candidate-pool size entering each level, narrowing from the full database to a single image. A green check marks the selection. The query shown is Vietnamese against an English database: \emph{g\'{\^a}u teddy Ikea} ranks the Djungelskog bear at 0.776, far above plush-toy distractors at roughly 0.4.}
  \label{fig:supp:retriever}
  \vspace{-3mm}
\end{figure*}

\begin{figure*}[t]
  \includegraphics[width=\textwidth]{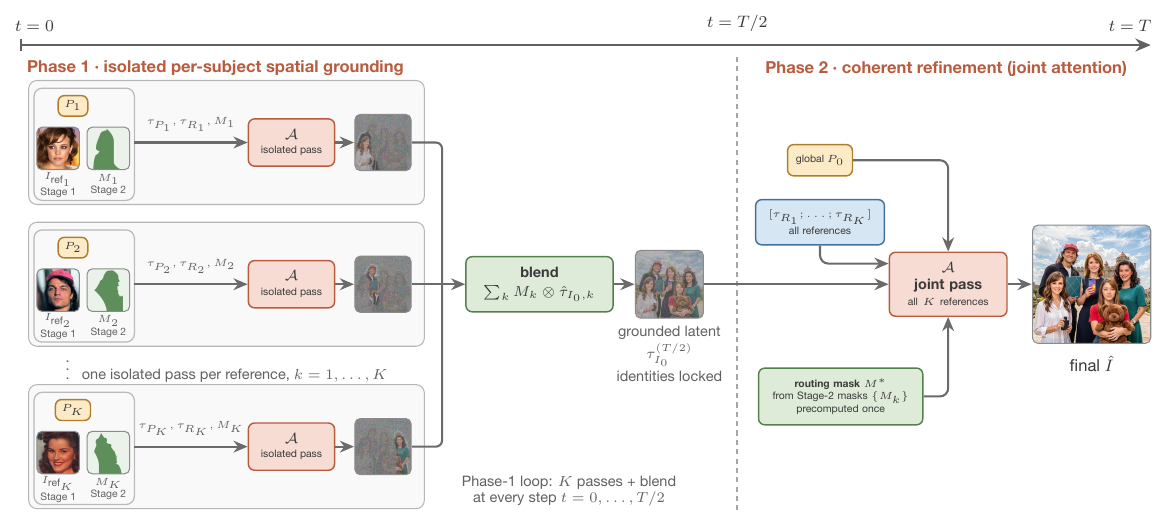}
  \vspace{-5mm}
  \caption{\textbf{Two-phase denoising.} Phase~1 denoises each reference in an isolated pass and mask-blends the results into the layout latent; Phase~2 denoises all references jointly under the global scene prompt and the spatial attention routing mask.}
  \label{fig:supp:two_phase}
  \vspace{-3mm}
\end{figure*}

\claude{The main paper takes the per-entity references as given (Sec.~4); this section describes the retriever that supplies them, with its three levels in full (\cref{fig:supp:retriever}).}

\noindent\textbf{Level~1: coarse text retrieval.} The entity name $r_k$ is encoded with BGE-M3~\cite{chen2024bgem3} and the top-$k_t$ database entities are taken by cosine similarity,
\begin{equation}
  \mathcal{C}_k^{(1)} = \operatorname{top\text{-}k_t}
  \bigl\{ \cos\bigl(\mathrm{BGE}(r_k),\, \mathrm{BGE}(r')\bigr) : r' \in \mathcal{D} \bigr\},
  \label{eq:supp:level1}
\end{equation}
where $\mathcal{D}$ is the set of entity names in the database. BGE-M3 is chosen for handling short, frequently non-English entity names without transliteration, which is what the worked query in \cref{fig:supp:retriever} exercises.

\noindent\textbf{Level~2: visual semantic filtering.} A level-1 entity carries multiple images across contexts and viewpoints. The pool narrows to the top-$k_v$ images most consistent with $c_k$ under CLIP-ViT-B/32~\cite{radford2021clip} cross-modal similarity,
\begin{equation}
  \mathcal{C}_k^{(2)} = \operatorname{top\text{-}k_v}
  \bigl\{ \cos\bigl(\mathrm{CLIP}_{\mathrm{img}}(I'),\, \mathrm{CLIP}_{\mathrm{txt}}(c_k)\bigr) : I' \in \mathcal{C}_k^{(1)} \bigr\}.
  \label{eq:supp:level2}
\end{equation}

\noindent\textbf{Level~3: fine-grained textual alignment.} The surviving $k_v$ candidates are scored by a cross-encoder over the concatenated (caption, context) pair,
\begin{equation}
  j^{*} = \underset{j \in \mathcal{C}_k^{(2)}}{\arg\max}\;
  \mathrm{Rerank}\bigl(\mathrm{caption}(I_j),\, c_k\bigr).
  \label{eq:supp:level3}
\end{equation}
Captions are produced offline by Qwen2.5-VL~\cite{bai2025qwen25vl} over object morphology, color, shape, and perspective. Generating them offline is what keeps MLLM latency out of the retrieval path; every reported number uses this pre-generated setting.

\subsection{Two-Phase Denoising}
\label{sec:supp:twophase}

\begin{figure*}[t]
    \centering
    \includegraphics[width=0.9\linewidth]{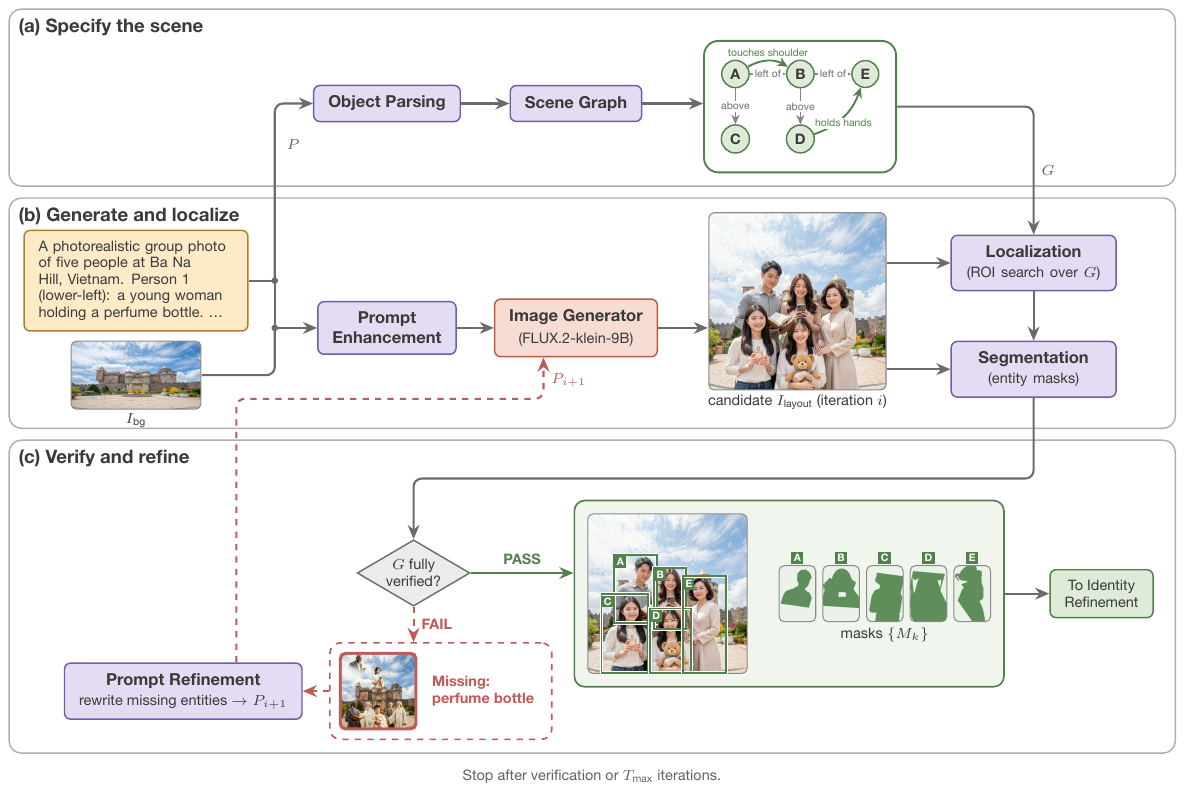}
    \vspace{-5mm}
    \caption{\textbf{Closed-loop layout generation and verification.} The generator produces a candidate layout $I_{layout}$ from the background and prompt. Localization and Segmentation agents check whether every entity in the scene graph $G$ is correctly instantiated. If any entity fails (dashed path), the Prompt Refinement agent revises the prompt and triggers a new cycle; the loop terminates when all entities pass (solid path).}
    \label{fig:supp:layout}
    \vspace{-3mm}
\end{figure*}

\claude{\cref{fig:supp:two_phase} details \revise{Stage~2 (Identity Refinement)} of the main paper (Eqs.~(2)--(5)). In Phase~1, $K$ isolated passes per step each carry one reference's tokens and are blended into the layout latent by their masks; in Phase~2, a single joint pass carries all references and the global prompt under the precomputed routing mask $M^{*}$.}

\subsection{Latent Encoding and Mask Preprocessing}
\label{sec:supp:encoding}

The $K$ binary masks are resampled onto the token grid rather than encoded. The path has to be exact, because a mask boundary disagreeing with the token grid puts the routing constraint \claude{of Eq.~(5)} on the wrong tokens. Each mask is first smoothed with a Gaussian blur (person and object kernels in \cref{tab:supp:hyperparams}) and then binarized at output resolution, so the blur only rounds the region boundary and leaves no soft values. The binary mask is then resized to the latent grid by nearest-neighbour interpolation and space-to-depth packed into the token layout, following the backbone's own latent preprocessing. Neither of these two steps averages, so the mask stays binary and its boundaries at token resolution coincide exactly with the regions transformer attention sees: the precondition both for the mask-weighted blending of Phase~1 and for the additive mask $M^{*}$.

Each reference latent additionally receives a dedicated 4D RoPE~\cite{su2024roformer} time-axis offset $\delta_k = k \cdot s_{\mathrm{scale}}$, placing references in disjoint slices of the rotary embedding space. Without it, two references at similar spatial positions in their own frames acquire incidental rotary proximity, and attention similarity stops reflecting semantic alignment alone.

\subsection{The Verification Loop}
\label{sec:supp:verifyloop}

\begin{algorithm}[t]
  \caption{Closed-loop layout generation and verification (\revise{Stage~1}).}
  \label{alg:supp:verify}
  \small
  \begin{algorithmic}[1]
    \Require prompt $P$, background $I_{bg}$, attribute descriptions $\{c_k\}$, threshold $\theta$, budget $T_{\max}$
    \State $G=(\mathcal{V},\mathcal{E}) \gets \textsc{SceneGraph}(\textsc{ObjectParsing}(P, I_{bg}))$
    \State $P_1 \gets \textsc{PromptEnhance}(P, \{c_k\})$;\; $\mathit{best} \gets \varnothing$
    \For{$i = 1, \ldots, T_{\max}$}
      \State $I_{layout} \gets \textsc{Generate}(I_{bg}, P_i)$;\; $F \gets \varnothing$
      \For{$v \in \mathcal{V}$ in parent-first order}
        \For{$R \in \mathrm{Anc}(v)$} \Comment{parent, ancestors, image}
          \State $(b_v, s) \gets \textsc{Localize}(I_{layout}, v, R)$
          \If{$s \ge \theta$} \textbf{break} \EndIf
        \EndFor
        \If{$s < \theta$} $F \gets F \cup \{(v, \pi(v), \textit{not detected})\}$; \textbf{continue} \EndIf
        \State $M_v \gets \textsc{Segment}(I_{layout}, b_v)$
        \If{$|M_v| = 0$} $F \gets F \cup \{(v, \pi(v), \textit{empty mask})\}$ \EndIf
      \EndFor
      \If{$|\mathcal{V}| - |F| >$ pass count of $\mathit{best}$} $\mathit{best} \gets (I_{layout}, \{M_v\})$ \EndIf
      \If{$F = \varnothing$} \Return $(I_{layout}, \{M_v\})$ \EndIf
      \State $P_{i+1} \gets \textsc{PromptRefine}(P_i, F)$
    \EndFor
    \State \Return $\mathit{best}$ \Comment{fallback after $T_{\max}$ iterations}
  \end{algorithmic}
\end{algorithm}

\cref{fig:supp:layout} illustrates the loop and \cref{alg:supp:verify} gives it in procedural form \claude{(Eq.~(1) of the main paper)}. Detections inside a region are assigned to scene-graph nodes by category-gated Hungarian matching~\cite{kuhn1955hungarian} with an IoU cost. The failure report $F$ is a list of (node, parent, reason) triples serialized as JSON and passed verbatim to the Prompt Refinement Agent, whose system prompt asks it to make the position and appearance of each listed node explicit while leaving the rest of $P_i$ unchanged. The check is node-level: adjacency and interaction edges of $G$ are not tested by a geometric predicate.


\subsection{Blockwise Routing Mask}
\label{sec:supp:maskblocks}

\begin{figure}[t]
    \centering
    \includegraphics[width=\linewidth]{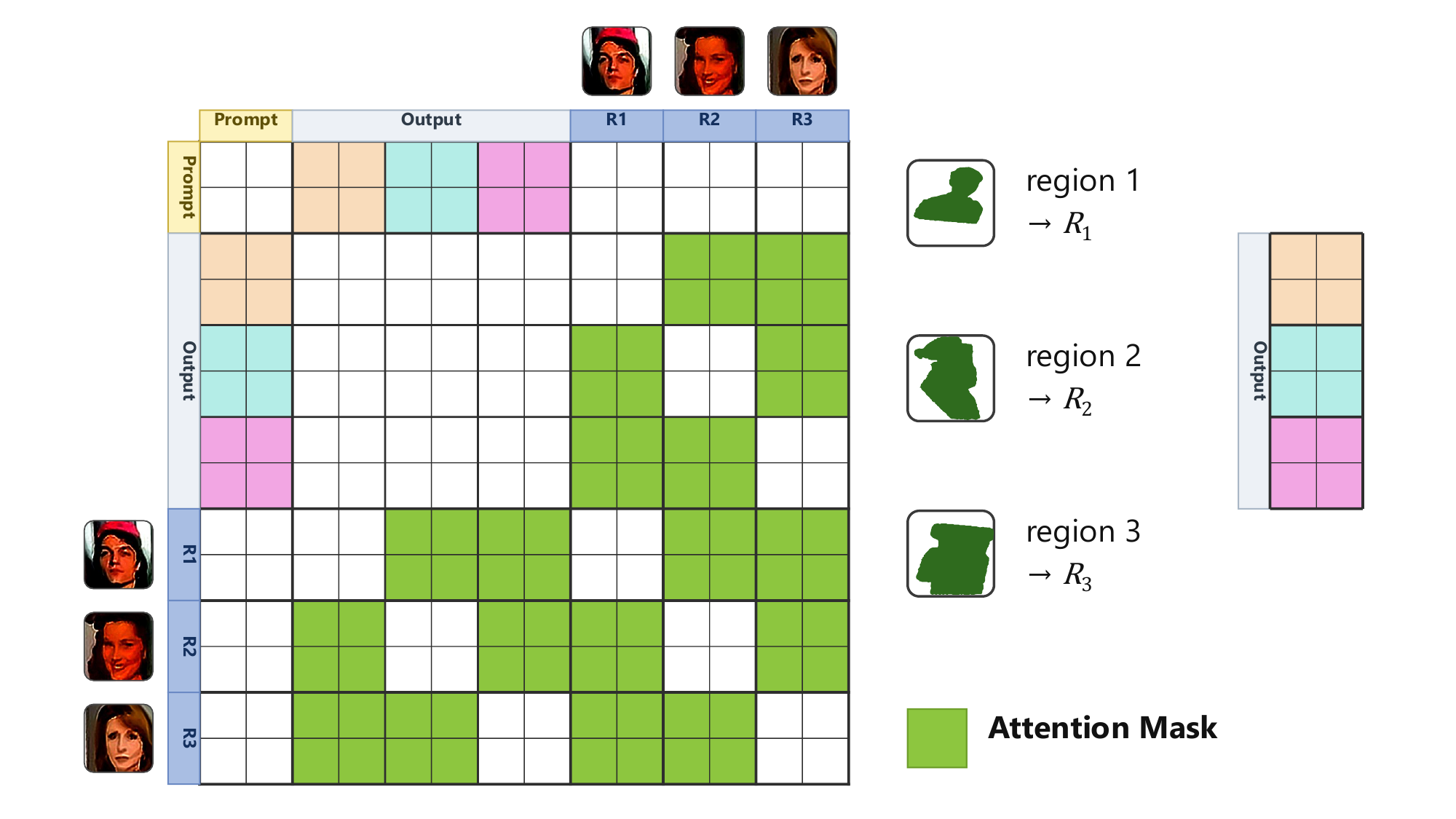}
    \vspace{-5mm}
  \caption{\textbf{Spatial attention routing mask $M^{*}$ for $K{=}3$.} Rows are queries and columns keys over the sequence [prompt\,|\,output\,|\,$R_1$\,|\,$R_2$\,|\,$R_3$]. Green cells are set to $-\infty$ before the softmax; white cells are unchanged. Output tokens see only their own reference, each reference sees only itself and its own region, and prompt tokens are unrestricted (Eq.~(5)).}
    \label{fig:supp:attnmask}
    \vspace{-3mm}
\end{figure}

\begin{table}[t]
  \centering
  \small
  \setlength{\tabcolsep}{4pt}
  \caption{\textbf{Blocks of the routing mask $M^{*}$} (\claude{Eq.~(5)}). Rows are query blocks, columns key blocks; ``open'' means the logit is unchanged, ``own'' means only the matching subject is open and every other entry is $-\infty$. $a(i)=0$ marks background output tokens.}
  \label{tab:supp:maskblocks}
  \begin{tabular}{@{}lccc@{}}
    \toprule
    \textbf{Query $\backslash$ Key} & \textbf{Text} & \textbf{Output} & \textbf{Ref.\ $R_{k'}$} \\
    \midrule
    Text                        & open & open & open \\
    Output, $a(i)=k \ge 1$      & open & open & own ($k'=k$) \\
    Output, $a(i)=0$            & open & open & blocked \\
    Reference $R_k$             & open & own ($a(j)=k$) & own ($k'=k$) \\
    \bottomrule
  \end{tabular}
\end{table}

\claude{\cref{fig:supp:attnmask} visualizes $M^{*}$ for $K{=}3$, and} \cref{tab:supp:maskblocks} spells out every block of $M^{*}$, so the mask can be rebuilt from the per-token assignment $a(\cdot)$ alone. Output-to-output and all text attention stay open, which leaves a path through which neighbouring subjects can still exchange information across layers; the main paper discusses what this does and does not guarantee.


\subsection{Attention Routing Diagnostics}
\label{sec:supp:wta}

\begin{figure}[t]
  \centering
  \includegraphics[width=\linewidth,trim={3.5cm 0 3cm 0},clip]{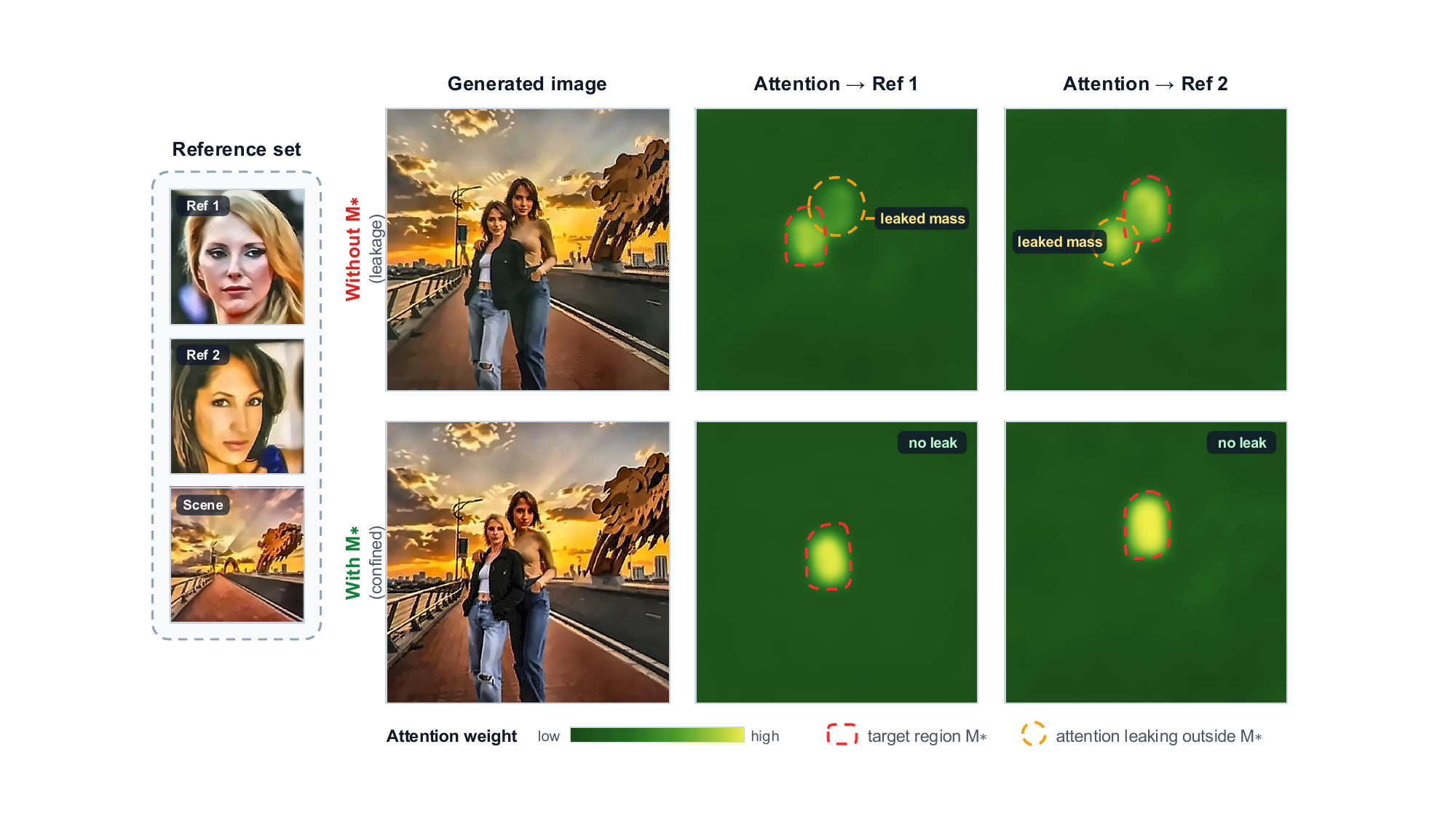}
  \vspace{-5mm}
  \caption{\textbf{Attention leakage under joint denoising, with and without $M^{*}$.} Columns: the generated image, then output-to-reference attention for Reference~1 and Reference~2 (brighter is higher). Top row without $M^{*}$, bottom row with it. Red outlines mark each reference's target region; dashed yellow outlines mark attention mass falling outside it. \claude{Without $M^{*}$ both references draw attention from outside their regions; with $M^{*}$ it stays inside.}}
  \label{fig:supp:attn_leakage}
  \vspace{-3mm}
\end{figure}

\claude{\cref{fig:supp:attn_leakage} shows per-reference attention with and without $M^{*}$, the evidence summarized in Sec.~4.2. Here we define the winner-take-all (WTA) map computed from the same attention, \revise{which turns that attention into a per-token routing label that can be inspected for any generated image}.} The construction is a diagnostic only: it reads out routing behaviour and never feeds back into generation.

At each denoising step $t$ and transformer block $l$ of the joint Phase-2 calls, the instrumented attention processors capture the output-to-reference sub-matrix
\begin{equation}
  A^{(l,t)}_{\mathrm{out}\to\mathrm{ref}}
  = \operatorname{softmax}\!\left(
      \frac{Q_{\mathrm{out}} K^{\top}}{\sqrt{d}} + M^{*}_{\mathrm{out\text{-}rows}}
    \right)_{:,\; T_{\mathrm{seq}}+O:},
  \label{eq:supp:attn}
\end{equation}
where $Q_{\mathrm{out}}$ are the queries of the $O$ output tokens and $M^{*}_{\mathrm{out\text{-}rows}}$ the output-token rows of the routing mask. Applying $M^{*}$ before the softmax makes the recorded scores the attention that was \emph{actually routed}. Averaging over the $C_k$ columns of reference $k$ gives
\begin{equation}
  s^{(l,t)}_{i,k} = \frac{1}{C_k} \sum_{j \in \mathcal{R}_k}
  A^{(l,t)}_{\mathrm{out}\to\mathrm{ref}}[i,\,j],
  \label{eq:supp:score}
\end{equation}
and each output token is assigned to its highest-scoring reference, $\mathrm{WTA}^{(l,t)}_i = \arg\max_k s^{(l,t)}_{i,k}$, aggregated over steps and blocks by majority vote,
\begin{equation}
  \mathrm{WTA}_i = \arg\max_{k} \sum_{l,t}
  \mathbf{1}\bigl[\mathrm{WTA}^{(l,t)}_i = k\bigr].
  \label{eq:supp:wta}
\end{equation}
Read against the input masks $\{M_k\}$, the map \revise{shows} disentanglement: winner regions that track the masks \revise{indicate that} the Phase-1 correspondence survived into Phase~2, and regions that do not localize where cross-reference interference persists. \revise{We use it as a qualitative diagnostic of individual cases, not as a benchmark metric; the quantitative effect of $M^{*}$ is the routing ablation of Tab.~4.}

\subsection{Agent Inventory}
\label{sec:supp:agents}

\begin{table}[t]
  \centering
  \small
  \setlength{\tabcolsep}{4pt}
  \caption{\textbf{The agents of \method, by stage.} Backing models are given by exact checkpoint in \cref{tab:supp:models}.}
  \label{tab:supp:agents}
  \begin{tabular}{@{}llp{2.4cm}@{}}
    \toprule
    \textbf{Agent} & \textbf{Model} & \textbf{Output} \\
    \midrule
    \multicolumn{3}{@{}l}{\textit{\revise{Reference Retrieval (optional; not used in the main experiments)}}} \\
    \quad Retrieval        & Qwen2.5-3B      & $(r_k, c_k)$, entity type \\
    \midrule
    \multicolumn{3}{@{}l}{\textit{\revise{Stage 1}: Layout Generation}} \\
    \quad Object Parsing   & Qwen2.5-VL-7B   & entities, containment \\
    \quad Scene Graph      & Qwen2.5-VL-7B   & graph $G$ \\
    \quad Prompt Enhance.  & Qwen2.5-VL-7B   & prompt $P_1$ \\
    \quad Image Generator  & FLUX.2-klein-9B & layout $I_{layout}$ \\
    \quad Localization     & Rex-Omni        & per-entity ROI \\
    \quad Segmentation     & SAM ViT-H       & masks $\{M_k\}$ \\
    \quad Prompt Refine.   & Qwen2.5-VL-7B   & prompt $P_{i+1}$ \\
    \midrule
    \multicolumn{3}{@{}l}{\textit{\revise{Stage 2}: Identity Refinement}} \\
    \quad (renderer)       & FLUX.2-klein-9B & image $\hat{I}$ \\
    \bottomrule
  \end{tabular}
\end{table}

\cref{tab:supp:agents} lists every agent, the model backing it, and what it emits. Four agents share one checkpoint: Object Parsing, Scene Graph, Prompt Enhancement, and Prompt Refinement are the same Qwen2.5-VL-7B-Instruct instance under different system prompts, which is why the loop's memory cost does not grow with the number of agent roles. The Identity Refinement stage has no agent of its own: it is the rendering backend invoked once verification passes.

\section{Implementation and Reproducibility}
\label{sec:supp:impl}

\subsection{Hyperparameters}
\label{sec:supp:hyperparams}

\begin{table}[t]
  \centering
  \small
  \setlength{\tabcolsep}{5pt}
  \caption{\textbf{\method hyperparameters, by stage.} Every value here was fixed for all reported runs. $^{*}$The released configuration passes an img2img strength of 1.5, which the scheduler clamps to the full denoising schedule, so the effective value is 1.0.}
  \label{tab:supp:hyperparams}
  \begin{tabular}{@{}llr@{}}
    \toprule
    \textbf{Stage} & \textbf{Hyperparameter} & \textbf{Value} \\
    \midrule
    Retrieval & Level-1 pool $k_t$ & 1 \\
              & Level-2 pool $k_v$ & 5 \\
    \midrule
    Layout  & localization confidence $\theta$ & 0.8 \\
            & max.\ loop iterations $T_{\max}$ & 3 \\
    \midrule
    Render  & output resolution & $1024^2$ \\
            & VAE compression $f$ & 8 \\
            & denoising steps $T$ & 16 \\
            & phase split & $T/2$ (step 8) \\
            & guidance scale & 1.5 \\
            & img2img strength & 1.0$^{*}$ \\
            & RoPE offset scale $s_{\mathrm{scale}}$ & 10 \\
            & face mask dilation & 2\,px \\
            & blur kernel (person / object) & 6 / 12 \\
    \bottomrule
  \end{tabular}
\end{table}

\cref{tab:supp:hyperparams} lists the fixed configuration. \revise{The phase split is fixed at the midpoint of the 16-step schedule, so that Phase~1 and Phase~2 receive equal shares of the denoising trajectory.} The localization threshold $\theta$ governs when the Localization Agent escalates from a parent ROI to the next ancestor, and $T_{\max}$ caps the verify-and-refine loop before the best partial layout is taken as a fallback. \claude{With $k_t{=}1$, Level~1 commits to a single entity and Levels~2--3 choose among that entity's images; Level~1 errors are therefore unrecoverable\revise{.}} Detection output is assigned to scene-graph entities by Hungarian matching with an IoU cost, gated so that only same-category pairs may match. This is a pipeline-side step; it is not how the evaluation metrics pair references to generated content (\cref{sec:supp:metrics}). \claude{The published baselines use their official Hugging Face checkpoints with the default sampling settings of each release; only the inputs (reference images, prompt, output resolution) are adapted to \bench, and none of them exposes a background-image channel.}

\subsection{Model Versions}
\label{sec:supp:models}

\begin{table}[t]
  \centering
  \small
  \setlength{\tabcolsep}{4pt}
  \caption{\textbf{Model versions and checkpoints.} Components marked $\dagger$ are shared between the generation pipeline and the evaluation suite. The Attr-VQA judge is independent of every pipeline model.}
  \label{tab:supp:models}
  \resizebox{\linewidth}{!}{%
  \begin{tabular}{@{}lp{4.3cm}@{}}
    \toprule
    \textbf{Component} & \textbf{Checkpoint} \\
    \midrule
    \multicolumn{2}{@{}l}{\textit{Generation pipeline}} \\
    \quad Backbone        & \texttt{black-forest-labs/}\newline\texttt{FLUX.2-klein-9B} \\
    \quad Layout agents   & \texttt{Qwen/Qwen2.5-VL-7B-}\newline\texttt{Instruct} (4-bit) \\
    \quad Retrieval agent & \texttt{Qwen/Qwen2.5-3B} \\
    \quad Detector        & \texttt{IDEA-Research/Rex-Omni} \\
    \quad Segmenter$^{\dagger}$ & SAM ViT-H \newline (\texttt{sam\_vit\_h\_4b8939.pth}) \\
    \quad Face parsing    & facexlib BiSeNet \\
    \quad Background removal & \texttt{BiRefNet} \\
    \midrule
    \multicolumn{2}{@{}l}{\textit{Benchmark construction}} \\
    \quad \claude{Attributes, prompts, verification} & \claude{\texttt{gemma-4-e4b-it}} \\
    \quad \claude{Deduplication / filtering} & \claude{DINOv2, LAION v2.5, CLIP ViT-L/14} \\
    \midrule
    \multicolumn{2}{@{}l}{\textit{Evaluation}} \\
    \quad ID-Sim          & ArcFace IR-SE50 \newline (\texttt{model\_ir\_se50.pth}) \\
    \quad DINO-Sim        & Florence-2-large + SAM$^{\dagger}$ \newline + DINOv2 \\
    \quad CLIP-T          & \texttt{openai/clip-vit-large-}\newline\texttt{patch14} \\
    \quad BG-Sim          & SAM~3 + DINOv3 \\
    \quad Aesth.          & \texttt{aesthetic\_predictor\_v2\_5} \\
    \quad Attr-VQA judge  & \texttt{gemma-4-e4b-it} \\
    \bottomrule
  \end{tabular}%
  }
\end{table}
\cref{tab:supp:models} pins every model in the pipeline, the benchmark construction, and the evaluation suite: FLUX.2-klein~\cite{bfl2025flux2klein}, Qwen2.5-VL~\cite{bai2025qwen25vl}, Qwen2.5~\cite{qwen2025qwen25}, Rex-Omni~\cite{jiang2026rexomni}, SAM~\cite{kirillov2023sam}, BiSeNet~\cite{yu2018bisenet}, BiRefNet~\cite{zheng2024birefnet}, Florence-2~\cite{xiao2024florence2}, SAM~3~\cite{carion2025sam3}, DINOv3~\cite{simeoni2025dinov3}, and Gemma~4~\cite{gemma2026gemma4}. \claude{The Attr-VQA judge shares no model with the generation pipeline, but it shares the Gemma~4 family with benchmark construction; the main paper lists this among the limitations.}

\subsection{Determinism, Seeds, and Compute}
\label{sec:supp:determinism}

Each benchmark case is assigned a fixed 32-bit seed derived as \texttt{zlib.adler32(case\_uid.encode())}, so noise realisations are identical across reruns. The flow-matching sampler uses a linear timestep schedule over 16 steps with the phase split at the midpoint, step~8. PyTorch deterministic mode is not enforced, and minor numerical differences across GPU architectures are possible without affecting aggregate metrics in our testing.

Two sources of non-determinism remain, both agentic rather than numerical. Qwen agents sample at temperature $0.7$, so successive runs produce slightly different rewrites, and the verification loop may consequently converge in a different number of iterations. The determinism claimed for routing is a property of the routing rule, not of the run. \revise{All tables report a single run per case and therefore do not measure this run-to-run variation; the conclusions of the main paper rest on margins of several points, larger than such variation could plausibly produce.}

\claude{\noindent\textbf{Compute.} Benchmark construction consumed about 38 GPU-hours on a workstation with $4\times$ A100 40\,GB: 12\,h for DINOv2 deduplication and CLIP filtering of about 15{,}000 candidate object/fashion images, 18\,h for prompt synthesis and verification of about 4{,}000 candidate prompts, and 8\,h of model time supporting the human-in-the-loop demographic audit; human annotation added roughly 60 person-hours across two annotators. Each published baseline and SRI generate the 1{,}361 images in 10--12\,h on the 4-GPU node. \method runs on two H100 80\,GB GPUs with zero-swap placement (backbone on GPU~0 during identity refinement, Qwen agents on GPU~1) at roughly 3--6 minutes per case, depending on subject count and loop iterations; the profiling figures of \cref{sec:supp:efficiency} instead come from a single A100 40\,GB with sequential offloading and are an unoptimized upper bound. Scoring one method on all six axes costs about 8 GPU-hours, of which Attr-VQA accounts for about 78\% because its query count grows as $O\bigl(N+\binom{N}{2}\bigr)$. No method is trained or fine-tuned, so reproducing every table requires only the released data, public checkpoints, and scripts.}

\section{Additional Quantitative Results}
\label{sec:supp:quant}

\subsection{Per-Group-Size Profiles}
\label{sec:supp:radar}

\begin{figure*}[t]
  \centering
  \includegraphics[width=\textwidth]{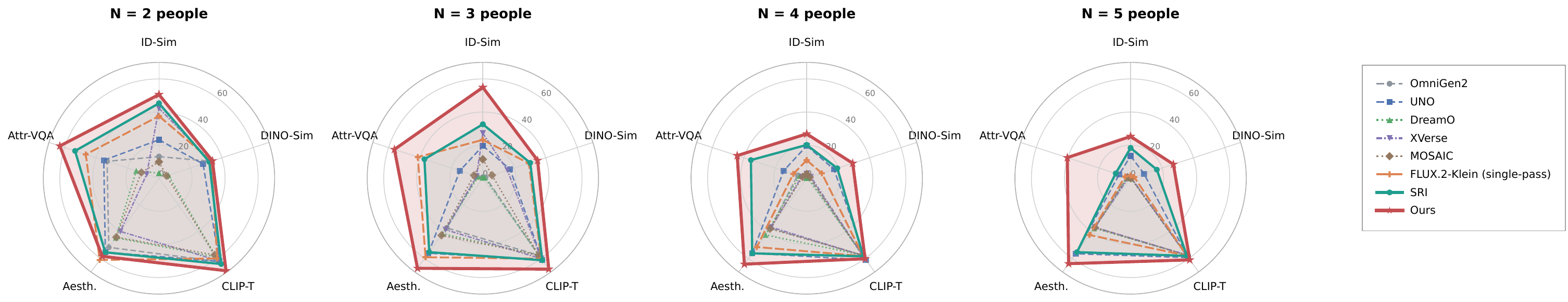}
  \vspace{-5mm}
  \caption{\textbf{Per-group-size profiles on \bench.} One panel per group size $N$, left to right; five axes clockwise from ID-Sim at the top; rings at 20, 40 and 60. All four panels share a single radial scale, so profiles are comparable across panels and further from the centre is better. Ours is solid red, SRI solid teal, and the same-backbone single-pass control dashed orange. BG-Sim is not plotted. \claude{Exact values are in Tab.~2.}}
  \label{fig:supp:radar}
  \vspace{-3mm}
\end{figure*}

\claude{\cref{fig:supp:radar} visualizes Tab.~2. The published methods' profiles contract toward the CLIP-T and Aesth.\ axes as $N$ grows, the visual form of the observation that global metrics stay flat while identity and binding collapse.}

\subsection{Efficiency}
\label{sec:supp:efficiency}

\begin{table}[t]
  \centering
  \small
  \caption{\textbf{Per-image efficiency against a single-pass FLUX generation.} Profiled on one A100 40\,GB with sequential offloading; ``---'' marks a stage the single-pass baseline does not have. \revise{The main paper's total (430.27\,s, $5.46\times$) excludes the optional Reference Retrieval.}}
  \label{tab:supp:efficiency}
  \begin{tabular}{@{}lrr@{}}
    \toprule
     & \textbf{\method} & \textbf{Single-pass} \\
    \midrule
    \multicolumn{3}{@{}l}{\textit{Latency (s/image) $\downarrow$}} \\
    \quad Reference Retrieval    & 35.22  & --- \\
    \quad Layout Generation      & 339.12 & --- \\
    \quad Identity Refinement    & 91.15  & 78.81 \\
    \quad \textbf{Total}         & \textbf{465.49} & \textbf{78.81} \\
    \midrule
    \multicolumn{3}{@{}l}{\textit{Peak GPU memory (GB) \revise{$\downarrow$}}} \\
    \quad Mean / median          & 38.7 / 38.7 & 22.1 / 21.9 \\
    \quad Maximum                & 39.3        & 24.7 \\
    \bottomrule
  \end{tabular}
\end{table}

\cref{tab:supp:efficiency} breaks \revise{the $5.46\times$ latency of the main paper down by stage; as there, the total excludes Reference Retrieval (430.27\,s against 78.81\,s)}. The dominant term is closed-loop layout verification at 339.12\,s, \revise{$78.8\%$} of the total, and the ablation of Tab.~4 shows that verification also contributes substantially to identity and binding. The latency is therefore the price of verification rather than of the generation mechanism. Two-phase denoising itself costs $+16\%$ over a single pass (91.15\,s against 78.81\,s), well under the $K\times$ its per-reference passes suggest: Phase~1 spans only the first $T/2$ of the 16 steps, each pass conditions on a single reference's short token sequence rather than the full $K$-reference concatenation, and the $K$ passes are batched on-device where VRAM permits. Memory peaks at 38.7\,GB on average against 22.1\,GB for the baseline, during identity refinement. Sequential offloading keeps the pipeline inside a single 40\,GB device, paying for multi-model orchestration in wall-clock time rather than in additional GPUs. \claude{Because verification dominates, so do the available savings: re-checking only the nodes that failed in the previous iteration, and caching detections across iterations; neither is implemented yet.}

\subsection{\texorpdfstring{\revise{Learned Routing, Canvas Conditioning, and Methods Not Compared}}{Learned Routing, Canvas Conditioning, and Methods Not Compared}}
\label{sec:supp:notcompared}

\revise{Several recent methods also decouple \emph{where} a reference acts from \emph{what} it contributes, but learn this decoupling. MOSAIC~\cite{mosaic2025} supervises reference--target correspondence and adds disentanglement losses so that subjects do not mix in attention. RefRoute~\cite{he2026refroute} compresses each reference into a short token sequence with a pixel-residual encoder and routes it to its target region by an attention mask, keeping latency growth low at 10--17 references after many-reference fine-tuning. RefCompose~\cite{kuppa2026refcompose} and Canvas-to-Image~\cite{dalva2025canvas} pack references and controls into one fixed-resolution canvas, so the conditioning cost stays constant in the number of references; RefCompose trains dual LoRA streams for structure and appearance, and Canvas-to-Image fine-tunes on layouts, poses, and subject insertion jointly. \method shares their goal of preventing cross-reference leakage but makes a different trade-off: it trains nothing, fixes each output token's routing from a layout that is verified before sampling, and keeps full-resolution reference tokens, paying instead with the latency of the verification loop (Sec.~5.4). Mask-based routing gives a hard, inspectable assignment without paired multi-subject training data, whereas learned routing and canvases scale to more references at lower cost and can learn interactions that a fixed mask cannot express. The two are complementary: a verified layout and its masks could supply the region assignment that learned routing consumes or the placement of a reference canvas.}

\revise{RefRoute~\cite{he2026refroute} and RefCompose~\cite{kuppa2026refcompose} are the recent methods closest to ours in mechanism: both route each reference to its own target region and separate where a reference acts from what it contributes (Sec.~2). Neither has publicly released its source code at the time of submission. Both also depend on training, many-reference fine-tuning for RefRoute and dual LoRA streams for RefCompose, so a reimplementation from the papers alone would not give a faithful comparison. We therefore discuss them in the related work but do not include them in Tab.~2. Because we release \bench, its evaluation scripts, and the SRI and \method pipelines, both methods can be evaluated under the same protocol once their code becomes available.}

\section{Human Preference Study}
\label{sec:supp:human}

\noindent\textbf{Protocol.} Each screen shows the prompt, the background and per-subject references, and two candidate images. For every pair the rater answers three forced-choice questions, covering overall preference, identity preservation, and prompt following, each with \textit{tie} and \textit{neither is good} options. Every pair puts \method against one baseline. We study the four baselines that produce competitive composites: UNO, OmniGen2, XVerse, and MOSAIC. DreamO is excluded because its outputs collapse to near-zero identity at every group size, which makes a forced choice uninformative. Gold attention-check pairs filter inattentive raters, who are removed before any analysis. After filtering, 12 raters contributed 972 dimension-level judgments over 324 comparisons. Per dimension we compute the head-to-head win rate against each baseline, tested with a two-sided sign test on decided (non-tie) pairs, and a global skill rating from a Bradley--Terry model~\cite{bradley1952rank} mapped onto the Elo scale, with $95\%$ confidence intervals from 2{,}000 bootstrap resamples. Inter-rater reliability uses Krippendorff's~$\alpha$~\cite{krippendorff2018content}.

\begin{figure}[t]
  \centering
  \includegraphics[trim={3em 0 0 0},clip,width=\linewidth]{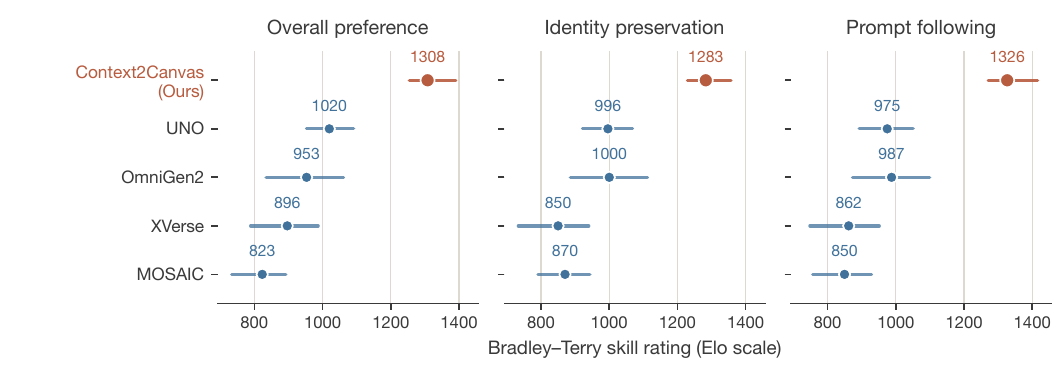}
  \vspace{-5mm}
  \caption{\textbf{Bradley--Terry ability on the Elo scale, per dimension.} Points are the estimate, bars the $95\%$ confidence interval from 2{,}000 bootstrap resamples. Our interval is disjoint from every baseline on all three dimensions; the baselines sit in a narrow band beneath, whose internal order is not stable across dimensions.}
  \label{fig:supp:elo}
  \vspace{-3mm}
\end{figure}

\begin{figure}[t]
  \centering
  \includegraphics[width=\linewidth,trim={1.5cm 0 0 0},clip]{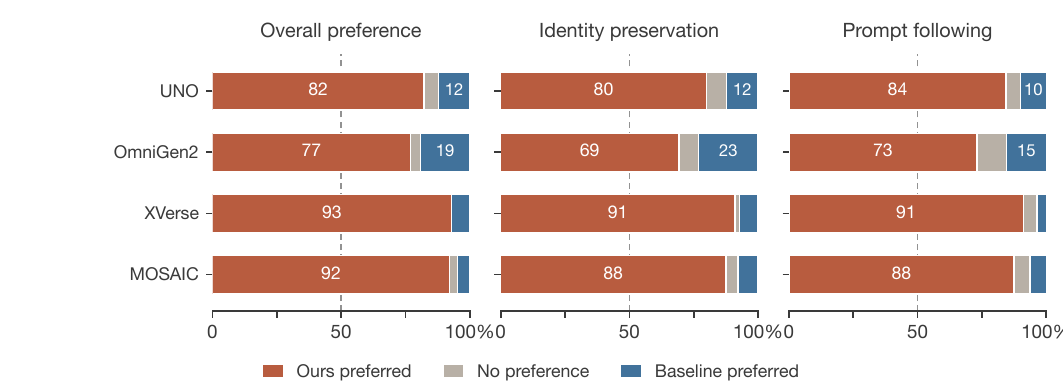}
  \vspace{-5mm}
  \caption{\textbf{Blind human head-to-head preference, per dimension.} Each bar splits the pairwise judgments into ours preferred, no clear preference, and baseline preferred; the dashed line marks the $50\%$ chance level.}
  \label{fig:supp:winrate}
  \vspace{-3mm}
\end{figure}

\noindent\textbf{Win rates and Bradley--Terry ratings.} \claude{\cref{fig:supp:winrate} gives the head-to-head preferences summarized in Sec.~5.6, and} \cref{fig:supp:elo} gives the skill ratings. \method ranks first on all three dimensions with a confidence interval disjoint from every baseline, while the baselines occupy a narrow band beneath it whose internal ordering is not stable across dimensions. Converting each Elo gap into an expected win probability reproduces the measured head-to-head rates to within a few percentage points, so the ranking is not an artifact of either estimator taken alone.

\noindent\textbf{Inter-rater reliability.} Krippendorff's $\alpha$ is low in absolute terms ($\approx 0.30$ overall) and lower still on the two component dimensions (0.18 on identity, 0.21 on prompt following). Absolute agreement of this order is expected when the axes are subjective and the comparison is forced, but it does not undermine the conclusion, for three independent reasons. The effect is large, ranging from $77\%$ to $93\%$ preference. The sign test is rater-agnostic: it counts decided pairs without assuming raters agree with one another, and every head-to-head clears $p<0.05$, with $p<10^{-10}$ for all but OmniGen2 at $p<0.01$. And the Bradley--Terry fit, which pools evidence differently again, returns confidence intervals disjoint from every baseline. The comparison against OmniGen2 is the closest on all three measures, and even there the preference is significant.

\section{Additional Qualitative Results}
\label{sec:supp:qual}

\subsection{Same-Backbone Control}
\label{sec:supp:samebackbone}

\begin{figure*}[t]
  \centering
  \includegraphics[width=\textwidth]{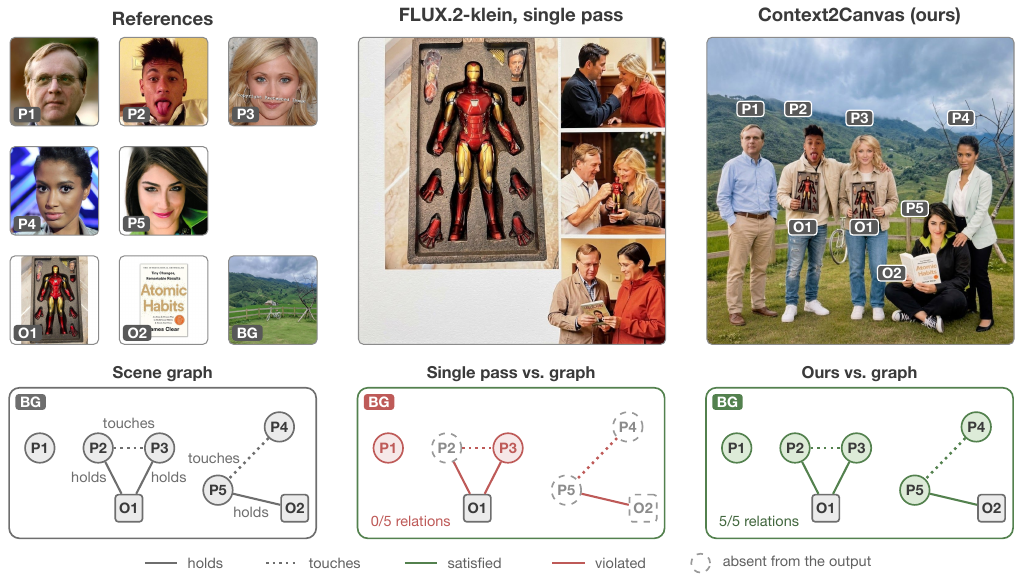}
  \vspace{-5mm}
  \caption{\textbf{Same-backbone control.} References and background (left), then naive single-pass FLUX.2-klein with all reference tokens concatenated into one call (middle), then full \method (right). The single-pass output leaks identities and misplaces attributes despite using the identical backbone.}
  \label{fig:supp:samebackbone}
  \vspace{-3mm}
\end{figure*}

\claude{The main paper argues numerically that the gain comes from the pipeline rather than the backbone (Tab.~2).} \cref{fig:supp:samebackbone} is the visual form of that argument. Both columns run FLUX.2-klein on identical references, background, and prompt; the only difference is whether retrieval, layout verification, and two-phase routing sit in front of it.

\subsection{Failure Cases}
\label{sec:supp:failures}

\begin{figure*}[t]
  \centering
  \includegraphics[width=\textwidth]{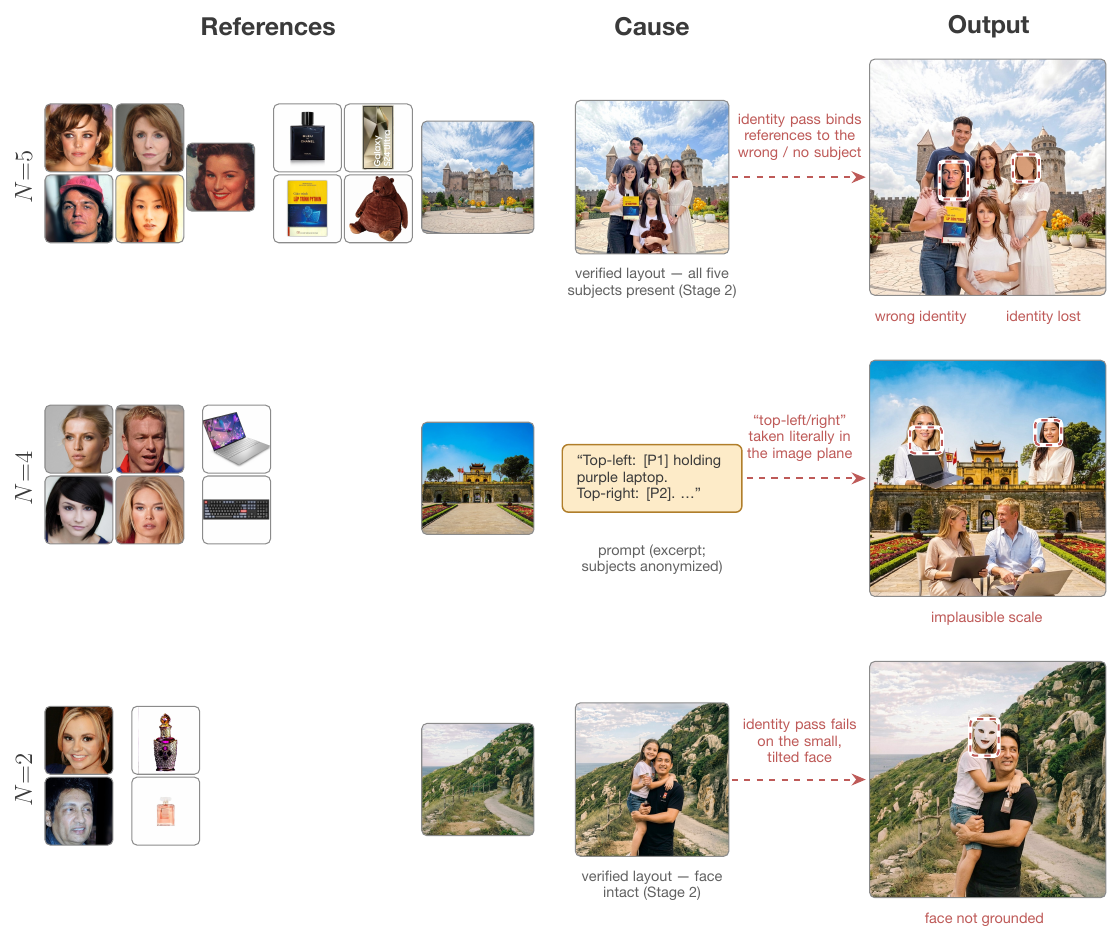}
  \vspace{-5mm}
  \caption{\textbf{Failure cases of \method.} Dashed red boxes mark the defects; the \emph{Cause} column shows the evidence isolating the failing step. Top and bottom: the verified intermediate layouts are correct, so the fault lies in the identity-grounding pass. Middle: the prompt's ``top-left'' and ``top-right'' are honored literally in the image plane, and verification passes the implausible layout because it checks presence and position, not plausibility. Prompt excerpts are verbatim with subject identifiers anonymized to [P1], [P2].}
  \label{fig:supp:failure}
  \vspace{-3mm}
\end{figure*}

\cref{fig:supp:failure} expands the two failures diagnosed in the main paper and adds a third. What makes them worth showing is that each isolates a different component, and the intermediate artifacts say which.

\noindent\textbf{Identity grounding degrades first as $N$ grows.} In the mountain-resort case at $N{=}5$, layout generation completed correctly: the verified base image contains all five subjects in front of the castle. The identity pass then fails twice, applying one reference to the wrong subject and leaving a second subject with a featureless face. The third case is an extreme-pose variant of the same failure, where a child carried in an adult's arms has a small, strongly tilted face that is localized correctly but rendered featureless. Both are consistent with the ID-Sim trend in \claude{Tab.~2}.

\noindent\textbf{Verification checks presence, not plausibility.} In the historic-citadel case the identities are correct, but layout generation satisfies the prompt's ``top-left'' and ``top-right'' literally in the image plane, placing two subjects at architecturally impossible scale. Verification passes the layout, because every entity in the scene graph is present \claude{and segmentable}. Nothing in the graph encodes physical plausibility, which is the limit of what scene-graph verification can catch.

These inputs are hard for every method: on the same mountain-resort prompt the single-pass control loses the castle and hallucinates a sixth person. The failures therefore concentrate at the identity-grounding and plausibility-checking frontiers rather than reflecting a general weakness in the pipeline.




\section{\texorpdfstring{\revise{Broader Impact}, Ethics, and Release}{Broader Impact, Ethics, and Release}}
\label{sec:supp:impact}

\noindent\textbf{Intended use.} \claude{\bench is a research benchmark and \method a research prototype for compositional, identity-preserving generation. Neither is offered as a deployable system for identity-conditioned generation, where consent and downstream use are not controlled.}

\revise{\noindent\textbf{Impact on evaluation practice.} The main finding of \bench is methodological: global scores stay nearly flat while identity, binding, and background fidelity collapse as subjects are added (Sec.~5.2). If adopted, its per-subject axes (ID-Sim and DINO-Sim under a stated assignment rule, BG-Sim, and Attr-VQA with graph-derived questions) would make this failure visible in routine evaluation, where CLIP-T and aesthetic scores currently hide it. Because Attr-VQA questions are derived from released graphs rather than written by hand, the protocol can be extended to new prompts and re-scored with other judges.

\noindent\textbf{Where the method fits.} \method trades speed for verifiable correctness: it is about $5.5\times$ slower than a single pass of its backbone (Sec.~5.4), and most of that cost is the verification loop. It therefore suits offline, correctness-critical uses in which a wrong identity or a swapped item is costly and a few minutes per image are acceptable, such as storyboards, previsualization, and constrained group compositions, rather than interactive editing. Every intermediate (scene graph, failure report, masks, and routing mask) can be inspected, so a user can see why an output is accepted or rejected.}

\noindent\textbf{Positive impacts.} The benchmark exposes systematic weaknesses in current multi-subject generators that are invisible to single-subject benchmarks, providing a target for methodological progress. Including Vietnamese landmarks (Ha Long Bay, Dragon Bridge, Fansipan, Ba Na Hills, Hoi An Old Town) in the background pool broadens cultural representation in compositional generation benchmarks, which to date have been dominated by Western scenes. \claude{Because both the benchmark protocol and the method are training-free and ship with evaluation scripts, they lower the barrier to entry for groups without large training budgets.}

\noindent\textbf{Risks and mitigations.} \claude{The salient risk of any identity-conditioned generator is non-consensual likeness, and a training-free method sharpens it: the cost of producing a new identity is a reference image rather than a training run. Four properties limit, without eliminating, what this work contributes to that risk. The identities are drawn from an established public research corpus of public figures~\cite{karras2018progressive}, and we add no new identities. We redistribute only a reference manifest (source links and structured metadata) and the prompt set; users obtain CelebA-HQ separately under its license, which forecloses one-click bulk download of the identity pool. The license is research-only and prohibits impersonation, defamation, harassment, and any application producing images of a real person without verifiable consent. Finally, \revise{the retrieval module} cannot ground an identity that is absent from its database; this mitigation does not extend to \revise{the layout and identity stages}, which accept any reference image, so a deployment that lets users upload photos would need consent verification and provenance watermarking, which this prototype does not provide.}

\noindent\textbf{Bias and representation.} The identity pool is balanced along gender, age group, and ethnicity by a human-in-the-loop audit, but residual bias remains: CelebA-HQ over-represents European and East Asian celebrities relative to global demographics, and our 100-identity selection is a best-effort rebalance within that source rather than a representative sample of any population. Object and fashion references are likewise biased toward items with strong web visibility, including a deliberate over-representation of culturally distinctive Vietnamese items (\eg, Ao Dai variants) to support the geographic diversification of the background pool. Benchmark scores should therefore be read as comparative measurements across methods, not as evidence of any method's behavior on under-represented identities or items.

\noindent\textbf{Memorization.} Because the identity pool overlaps with generator pretraining corpora, ID-Sim and DINO-Sim partly reflect memorization rather than reference grounding. This affects every method compared here, so the relative ordering stands, but no absolute identity score should be read as evidence of generalization to identities a backbone has never seen. \bench is a high-fidelity testbed for real-world compositional generation, not a probe of out-of-distribution generalization.

\noindent\textbf{Reliance on learned judges.} \claude{Two of the six axes rest on learned judges: Attr-VQA on Gemma~4 and BG-Sim on SAM~3 with DINOv3. Neither is used in the generation pipeline, but Gemma~4 also wrote and verified the benchmark prompts, which couples the benchmark to one model family's blind spots. We release the full prompt and answer logs so that third parties can re-score the benchmark with alternative judges.}

\noindent\textbf{Release.} \claude{We release the prompt set, structured interaction and position graphs, reference metadata, the manifest pointing into CelebA-HQ and the original web sources, the evaluation scripts, and the \method and SRI pipelines, under a research-only license with the prohibitions above\revise{, together with a datasheet that documents the demographic composition of the identity pool (counts by gender, age group, and ethnicity), the source and license of every reference, and the known coverage gaps discussed under \emph{Bias and representation}}. 
Branded object and fashion imagery is referenced by source link rather than redistributed, so link rot may reduce long-term reproducibility; known issues will be tracked at the project page.} \revise{The benchmark dataset and source code will be publicly released upon paper acceptance.}



\end{document}